\documentclass[10pt,twocolumn,letterpaper]{article}

\usepackage{cvpr}
\usepackage{times}
\usepackage{epsfig}
\usepackage{graphicx}
\usepackage{amsmath}
\usepackage{amssymb}
\usepackage{booktabs}
\usepackage{multirow}
\usepackage{subfigure}

\usepackage[pagebackref=true,breaklinks=true,letterpaper=true,colorlinks,bookmarks=false]{hyperref}

\cvprfinalcopy 

\def\cvprPaperID{6115} 

\ifcvprfinal\fi
\begin{document}
\title{Unsupervised Deep Epipolar Flow for Stationary or Dynamic Scenes}

\author{Yiran Zhong$^{1,4,5}$, Pan Ji$^{2}$, Jianyuan Wang$^{1}$, Yuchao Dai$^{3}$, Hongdong Li$^{1,4}$\\
$^{1}$Australian National University, 
$^{2}$NEC Labs America, \\
$^{3}$Northwestern Polytechnical University,
$^{4}$ACRV, $^{5}$Data61 CSIRO\\
\tt\small{\{yiran.zhong, hongdong.li\}@anu.edu.au}, panji@nec-labs.com, daiyuchao@nwpu.edu.cn}

\maketitle
\thispagestyle{empty}

\begin{abstract}
Unsupervised deep learning for optical flow computation has achieved promising results.  Most existing deep-net based methods rely on image brightness consistency and local smoothness constraint to train the networks.  Their performance degrades at regions where repetitive textures or occlusions occur.  In this paper, we propose Deep Epipolar Flow, an unsupervised optical flow method which incorporates global geometric constraints into network learning. In particular, we investigate multiple ways of enforcing the epipolar constraint in flow estimation.  To alleviate a ``chicken-and-egg'' type of problem encountered in dynamic scenes where multiple motions may be present, we propose a low-rank constraint as well as a union-of-subspaces constraint for training. Experimental results on various benchmarking datasets show that our method achieves competitive performance compared with supervised methods and outperforms state-of-the-art unsupervised deep-learning methods.
\end{abstract}

\section{Introduction}
Optical flow estimation is a fundamental problem in computer vision with many applications. Since Horn and Schunck's seminal work~\cite{horn1981determining}, various methods have been developed using variational optimization~\cite{aubert1999computing,zach2007duality,brox2004high}, energy minimization~\cite{Kolmogorov2001,menze2015discrete,sun2010secrets,yang2015dense}, or deep learning~\cite{Dosovitskiy:2015,Ilg17,DBLP:journals/corr/RanjanB16,Sun2018PWC-Net}. In this paper, we particularly tackle the problem of unsupervised optical flow learning using deep convolutional neural networks (CNNs). Compared to its supervised counterpart, unsupervised flow learning does not require ground-truth flow, which is often hard to obtain, as supervision and can thus be applied in broader domains.

Recent research has been focused on transforming traditional domain knowledge of optical flow into deep learning, in terms of either training loss formulation or network architecture design. For example, in view of brightness consistency between two consecutive images, a constraint that has been commonly used in conventional optical flow methods, researchers have formulated photometric loss~\cite{Yu2016,Simon2018}, with the help of fully differentiable image warping~\cite{jaderberg2015spatial}, to train deep neural networks. Other common techniques including image pyramid~\cite{bouguet2001pyramidal} (to handle large flow displacements), total variation regularization~\cite{rudin1992nonlinear,wedel2009improved} and occlusion handling~\cite{alvarez2007symmetrical} have also led to either new network structures (\eg, pyramid networks~\cite{DBLP:journals/corr/RanjanB16,Sun2018PWC-Net}) or losses (\eg, smoothness loss and occlusion mask~\cite{Wang_2018_CVPR,Janai2018ECCV}). In the unsupervised setting, existing methods mainly rely on the photometric loss and flow smoothness loss to train deep CNNs. This, however, poses challenges for the neural networks to learn optical flow accurately in regions with repetitive textures and occlusions. Although some methods~\cite{Wang_2018_CVPR, Janai2018ECCV} jointly learn occlusion masks, these masks do not mean to provide more constraints but only to remove the outliers in the losses. In light of the difficulties of learning accurate flow in these regions, we propose to incorporate {\it global} epipolar constraints into flow network training in this paper.

Leveraging epipolar geometry in flow learning, however, is not a trivial task. An inaccurate or wrong estimate of fundamental matrices~\cite{hartley2003multiple} would mislead the flow network training in a holistic way, and thus significantly degrade the model prediction accuracy. This is especially true when a scene contains multiple independent moving objects as one fundamental matrix can only describe the epipolar geometry of one rigid motion. Instead of posing a hard epipolar constraint, in this paper, we propose to use soft epipolar constraints that are derived using {\it low-rankness} when the scene is stationary, and {\it union of subspaces structure} when the scene is motion agnostic. We thus formulate corresponding losses to train our flow networks unsupervisedly. 

Our work makes an attempt towards incorporating epipolar geometry into deep unsupervised optical flow computation.  Through extensive evaluations on standard datasets, we show that our method achieves competitive performance compared with supervised methods, and outperforms existing unsupervised methods by a clear margin. Specifically, as of the date of paper submission, on KITTI and MPI Sintel benchmarks, our method achieves the best performance among published deep unsupervised optical flow methods.

\section{Related work}
Optical flow estimation has been extensively studied for decades. A significant number of papers have been published in this area. Below we only discuss a few geometry-aware methods and recent deep-learning based methods that we consider closely related to our method.

\noindent{\bf Supervised deep optical flow.}
Recently, end-to-end learning based deep optical flow approaches have shown their superiority in learning optical flow. Given a large amount of training samples, optical flow estimation is formulated to learn the regression between image pair and corresponding optical flow. These approaches achieve comparable performance to state-of-the-art conventional methods on several benchmarks while being significantly faster. 
FlowNet \cite{Dosovitskiy:2015} is a pioneer in this direction, which needs a large-size synthetic dataset to supervise network learning. FlowNet2 \cite{Ilg17} greatly extends FlowNet by stacking multiple encoder-decoder networks one after the other, which could achieve a comparable result to conventional methods on various benchmarks. Recently, PWC-Net \cite{Sun2018PWC-Net} combines sophisticated conventional strategies such as pyramid, warping and cost volume into network design and set the state-of-the-art performance on KITTI \cite{Geiger2012CVPR,Menze2015CVPR} and MPI Sintel \cite{Butler:ECCV:2012}. These supervised deep optical flow methods are hampered by the need for large-scale training data with ground truth optical flow, which also limits their generalization ability.

\vspace{1mm}
\noindent{\bf Unsupervised deep optical flow.}
Instead of using ground truth flow as supervision, Yu \etal \cite{Yu2016} and Ren \etal \cite{ren2017unsupervised} suggested that, similar to conventional methods, the image warping loss can be used as supervision signals in learning optical flow. However, there is a significant performance gap between their work and the conventional ones. Then, Simon \etal \cite{Simon2018} analyzed the problem and introduced bidirectional Census loss to handle illumination variation between frames robustly. Concurrently, Yang \etal \cite{Wang_2018_CVPR} proposed an occlusion-aware warping loss to exclude occluded points in error computation. Very recently, Janai \etal \cite{Janai2018ECCV} extended two-view optical flow to multi-view cases with improved occlusion handling performance. Introducing sophisticated occlusion estimation and warping loss reduces the performance gap between conventional methods and current unsupervised ones, nevertheless, the gap is still huge. To address this issue, we propose a global epipolar constraint in flow estimation that largely narrows the gap.

\noindent{\bf Geometry-aware optical flow.}
In the field of cooperating with geometry constrains, Valgaerts \etal \cite{Valgaerts:2008} introduced a variational model to simultaneously estimate the fundamental matrix and the optical flow. Wedel \etal \cite{Wedel09} utilized fundamental matrix prior as a weak constraint in a variational framework. Yamaguchi \etal \cite{Yamaguchi13} converted optical flow estimation task into a 1D search problem by using precomputed fundamental matrices and the small motion assumptions. These methods, however, assume that the scene is mostly rigid (and thus a single fundamental matrix is sufficient to constrain two-view geometry), and treat the dynamic parts as outliers \cite{Wedel09}. Garg \etal \cite{Garg2010} used the subspace constraint on multi-frame optical flow estimation as a regularization term. However, this approach, assumes an affine camera model and works over entire sequences. Wulff \etal \cite{Wulff:CVPR:2017} used semantic information to split the scene into dynamic objects and static background and only applied strong geometric constraints on the static background. Recently, inspired by multi-task learning, people started to jointly estimate depth, camera poses and optical flow in an unified framework \cite{ranjan2018adversarial,yin2018geonet,zou2018df}. These work mainly leverage a consistency between flows that estimated from a flow network and computed from poses and depth. This constraint only works for stationary scenes and their performance is only comparable with unsupervised deep optical flow methods.

By contrast, our proposed method is able to handle both stationary and dynamic scenarios without explicitly computing fundamental matrices. This is achieved by introducing soft epipolar constraints derived from epipolar geometry, using {\it low-rankness} and {\it union-of-subspaces} properties. Converting these constraints to proper losses, we can exert global geometric constraints in optical flow learning and obtain much better performance.

\section{Epipolar Constraints in Optical Flow}
Optical flow aims at finding dense correspondences between two consecutive frames. Formally, let $I^t$ denote the image at time $t$, and $I^{t+1}$ the next image. For pixels $\mathbf{x}_i$ in $I^t$, we would like to find their correspondences $\mathbf{x}'_i$ in $I^{t+1}$. The displacement vectors $\mathbf{v} = [\mathbf{v}_1,...,\mathbf{v}_N]\in \mathbb{R}^{2\times N}$ (with $N$ the total number of pixels in $I^t$) are the optical flow we would like to estimate.

Recall that in two-view epipolar geometry~\cite{hartley2003multiple}, by using the homogeneous coordinates, a pair of point correspondences in two frames $\mathbf{x}'_i=(x_i',y'_i,1)^T$ and $\mathbf{x}_i=(x_i,y_i,1)^T$ is related by a fundamental matrix $\mathbf{F}$, 
\begin{equation}
\textbf{x}'^T_i\mathbf{F}\textbf{x}_i=0\;.
\label{eq:epi}
\end{equation}

In the following sections, we show how to enforce the epipolar constraint as a global regularizer in flow learning.

\subsection{Two-view Geometric Constraint}
Given estimated optical flow $\mathbf{v}$, we can convert it to a series of correspondences $\mathbf{x}_i$ and $\mathbf{x}'_i$ in $I^t$ and $I^{t+1}$ respectively. Then these corresponding points can be used to compute a fundamental matrix $\mathbf{F}$ by normalized 8 points method~\cite{hartley2003multiple}. Once the $\mathbf{F}$ is estimated, we can compute its fitting error. Directly optimizing Eq.~\eqref{eq:epi} is not effective as it is only an algebraic error that does not reflect the real geometric distances. We can use the Gold Standard method \cite{hartley2003multiple} to compute the geometric distances but it requires reconstructing the 3D points $\widehat{\mathbf{X}}_i$ beforehand for every point. Otherwise, we can use its first-order approximation, the Sampson distance $\mathcal{L}_{\text{\bf F}}$ to represent the geometric error,
\begin{equation}
\mathcal{L}_{\text{\bf F}} = \sum_i^N\frac{(\textbf{x}_i'^T\mathbf{F}\textbf{x}_i)^2}{(\mathbf{F}\textbf{x}_i)_1^2+(\mathbf{F}\textbf{x}_i)_2^2+(\mathbf{F}^T\textbf{x}'_i)_1^2+(\mathbf{F}^T\textbf{x}'_i)_2^2}\;.
\label{eq:fitF}
\end{equation}

The difficulty of optimizing this equation comes from its \emph{chicken-and-egg} character: it consists of two mutually interlocked sub-problems, \ie, \emph{estimating a fundamental matrix $\mathbf{F}$ from an estimated flow} and \emph{updating the flow to comply with the $\mathbf{F}$}. This alternating method, therefore, heavily relies on proper initialization.

Up to now, we have only considered the static scene scenario, where only ego-motion exists. In a multi-motion scene, this method requires estimating $\mathbf{F}$ for each motion, which again needs a motion segmentation step. It is still feasible to address this problem via iteratively solving three sub-tasks: (i) {\it update flow estimation}; (ii) {\it estimate $\mathbf{F}^m$ for each rigid motion given current motion segmentation}; (iii) {\it update motion segmentation based on the nearest $\mathbf{F}^m$}.

However, this method again has several inherent limitations. First, the number of motions need to be known as a priori which is almost impossible in general optical flow estimation. Second, this method is still sensitive to the quality of initial optical flow estimation and motion labels. Incorrectly estimated flow may generate wrong $\mathbf{F}^m$, which will in turn lead flow estimation to the wrong solution, therefore making the estimation even worse. Third, the motion segmentation step is non-differentiable, so with it, an end-to-end learning becomes impossible.

To overcome these drawbacks, we formulate two soft epipolar constraints using {\it low-rankness} and {\it union-of-subspaces} properties. And we will show that these constraints can be easily included as extra losses to regularize the network learning.

\subsection{Low-rank Constraint}
In this section, we show that it is possible to enforce a soft epipolar constraint without explicitly computing the fundamental matrix in a static scene.

Note that we can rewrite the epipolar constraint in Eq.~\eqref{eq:epi} as
\begin{equation}
\mathbf{f}^T\mathrm{vec}(\textbf{x}_i'\textbf{x}_i^T) = 0\;,
\end{equation}
where $\mathbf{f}\in \mathbb{R}^9$ is the vectorized fundamental matrix of $\mathbf{F}$ and 
\begin{equation}
\mathrm{vec}(\textbf{x}'_i\textbf{x}_i^T) = (x_ix'_i,x_iy'_i,x_i,y_ix'_i,y_iy'_i,y_i,x'_i,y'_i,1)^T\;.
\label{epipolar_subspace}
\end{equation}

Observe that, $\mathrm{vec}(\textbf{x}'_i\textbf{x}_i^T)$ lies on a subspace (of dimension up to eight), called epipolar subspace~\cite{ji2016robust}. Let us define $\mathbf{h}_i = \mathrm{vec}(\textbf{x}_i'\textbf{x}_i^T)$. Then the data matrix $\mathbf{H} = [\mathbf{h}_1,...,\mathbf{h}_N]$ should be low-rank. This provides a possible way of regularizing optical flow estimation via rank minimization instead of explicitly computing ${\bf F}$. Specifically, we can formulate a loss as
\begin{equation}
\mathcal{L}_{\text{lowrank}} = {\rm rank}(\mathbf{H})\;,
\label{eq:lowrank0}
\end{equation}
which is unfortunately non-differentiable and is thus not feasible to serve as a loss for flow network training. Fortunately, we can still use its convex surrogate, the nuclear norm, to form a loss as
\begin{equation}
\mathcal{L}_{\text{lowrank}}^* = \|\mathbf{H}\|_*\;,
\label{eq:lowrank1}
\end{equation}
where the nuclear norm $||\cdot||_*$ can be computed by performing singular value decomposition (SVD) of $\mathbf{H}$. Note that the SVD operation is differentiable and has been implemented in modern deep learning toolboxes such as Tensorflow and Pytorch, so this nuclear norm loss can be easily incorporated into network training. We also note that though this low-rank constraint is derived from epipolar geometry described by a fundamental matrix, it still applies in degenerate cases where a fundamental matrix does not exist. For example, when the motion is all zero or pure rotational, or the scene is fully planar, ${\bf H}$ will have rank six; under certain special motions, \eg, an object moving parallel to the image plane, its ${\bf H}$ will have rank seven.

Comparing to the original epipolar constraint, one may concern that this low-rank constraint is too loose to be effective, especially when the ambient space dimension is only nine. Although a thorough theoretical analysis is out of the scope of this paper (interested readers may refer to literature such as~\cite{recht2008necessary}), we will show in our experiments that this loss can improve the model performance by a significant margin when trained on data with mostly static scenes. However, this loss becomes ineffective when a scene has more than one motion, as the matrix ${\bf H}$ will then be full-rank.

\subsection{Union-of-Subspaces Constraint}
\begin{figure*}
    \centering
    \includegraphics[height=0.24\linewidth]{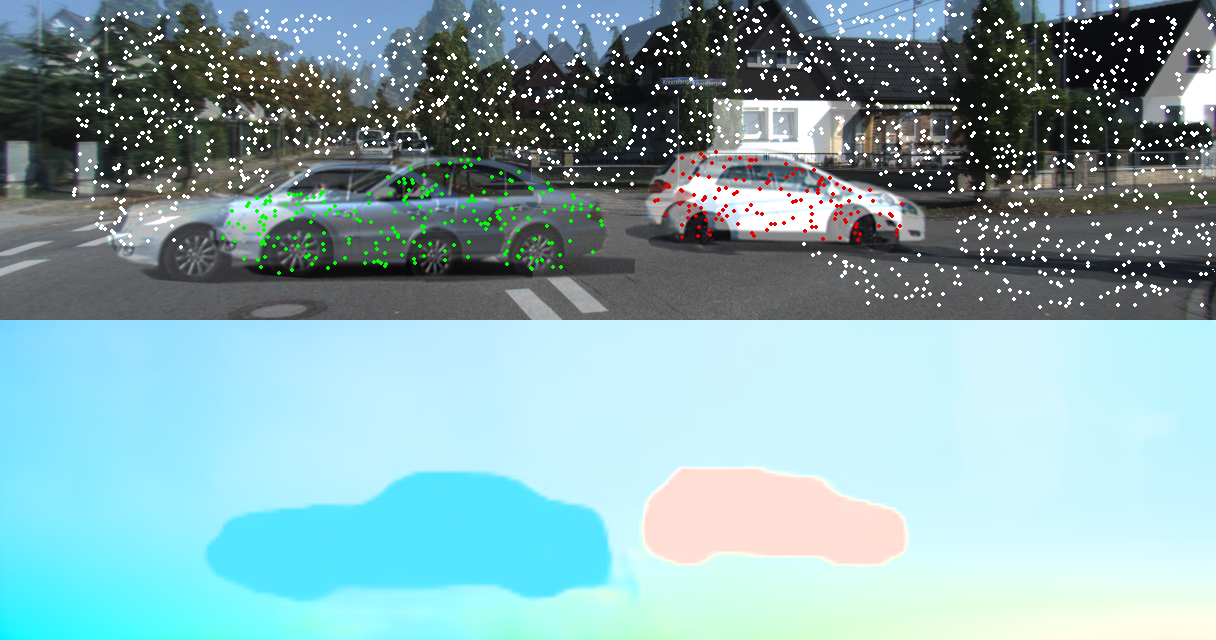} \hspace{0.4cm}
    \includegraphics[width=0.26\linewidth,height=0.24\linewidth]{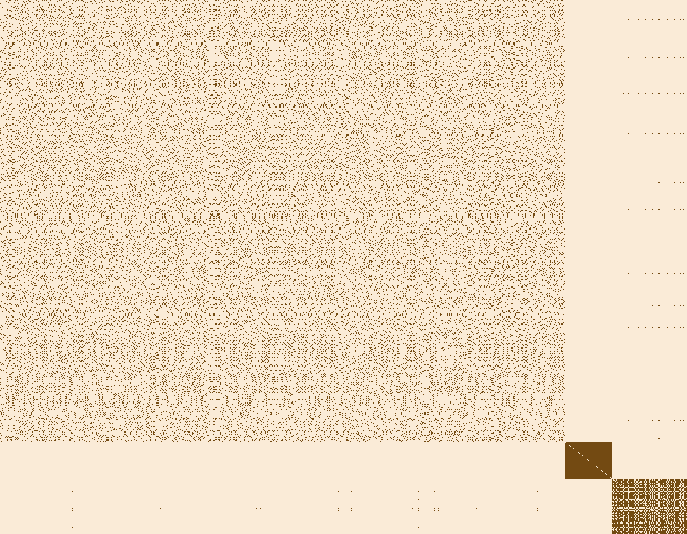}    
    \caption{\textbf{Motion segmentation and affinity matrix (constructed from ${\bf C}$) visualization.} The scene contains three motions annotated by three different colors: the ego-motion and the two cars' movements. On the right, we show a constructed affinity matrix from ${\bf C}$ which contains three diagonal blocks corresponding to these three motions. On the bottom left, we illustrate our estimated optical flow and the top left image shows that all these three motions are correctly segmented based on the ${\bf C}$. The sparse dots on the image are the sampled 2000 points that has been used to compute ${\bf C}$. It proves that our Union-of-Subspace constraint can work under multi-body scenarios.}
    \label{fig:examC}
    \vspace{-0.3cm}
\end{figure*}

In this section, we introduce another soft epipolar constraint, namely {\it union-of-subspaces constraint}, which can be applied in broader cases.

From Eq.~\eqref{epipolar_subspace}, it's not hard to observe that $\mathbf{h}_i$ from one rigid motion lies on a common epipolar subspace because they all share the same fundamental matrix. When there are multiple motions in a scene, $\mathbf{h}_i$ will lie in a union of subspaces. Note that this union-of-subspace structure has been shown to be useful in motion segmentation from two perspective images~\cite{li2013perspective}. Here, we re-formulate it in optical flow learning and come up with an effective loss using closed-form solutions.

In particular, the union-of-subspaces structure can be characterized by the self-expressiveness property~\cite{elhamifar2013sparse}, \ie, a data point in one subspace can be represented by a linear combination of other points from the same subspace. This has translated into a mathematical optimization problem~\cite{lu2012robust,ji2014efficient} as
\begin{equation}
\min\limits_{\bf C} {\frac{1}{2}}\Vert {\bf C}\Vert_{F}^{2} \quad \text{s.t.} \quad {\bf H}={\bf HC}\;.
\label{eq:noisefree}
\end{equation}
where ${\bf C}$ is the subspace self-expression coefficient and $\mathbf{H}$ is a matrix function of estimated flows. Note that, in subspace clustering literature, other norms on ${\bf C}$ have also been used, \eg, nuclear norm in~\cite{liu2013robust} and $\ell_1$ norm in~\cite{elhamifar2013sparse}. We are particularly interested in the Frobenius norm regularization due to its simplicity and equivalence to nuclear norm optimization~\cite{ji2014efficient}, which is crucial for formulating an effective loss for CNN training.

However, in real world scenarios, the flow estimation inevitably contains noises. Therefore, we relax the constraints in Eq.\eqref{eq:noisefree} by alternatively optimizing the function below
\begin{equation}
\mathcal{L}_{\text{subspace}} = {\frac{1}{2}}\Vert {\bf C}\Vert_{F}^{2}+{\frac{\lambda}{2}}\Vert {\bf HC}-{\bf H}\Vert_{F}^{2}\;,
\label{eq:sub}
\end{equation}
Instead of using a iterative solver, given an $\mathbf{H}$, we can derive a closed form solution for $\mathbf{C}$, \ie, 
\begin{equation}
{\bf C}^*=({\bf I}+\lambda {\bf H}^{T}{\bf H})^{-1}\lambda {\bf H}^{T}{\bf H}.
\end{equation}

Plugging the solution of ${\bf C}$ back to Eq.\eqref{eq:sub}, we arrive at our final union-of-subspaces loss term that only depends on the estimated flow:
\begin{equation}
\begin{split}
\mathcal{L}_{\text{subspace}} = &{\frac{1}{2}}\Vert ({\bf I}+\lambda {\bf H}^{T}{\bf H})^{-1}\lambda {\bf H}^{T}{\bf H}\Vert_{F}^{2}  \\+ &{\frac{\lambda}{2}}\Vert {\bf H}({\bf I}+\lambda {\bf H}^{T}{\bf H})^{-1}\lambda {\bf H}^{T}{\bf H}-{\bf H}\Vert_{F}^{2}\;.
\end{split}
\label{eq:sub1}
\end{equation}

Directly applying this loss to the whole image will lead to GPU memory overflow due to the computation of ${\bf H}^{T}{\bf H}\in \mathbb{R}^{N\times N}$ (with $N$ the number of pixels in a image). To avoid this, we employ a randomly sampling strategy to sample 2000 flow points in a flow map and compute a loss based on these samples. This strategy is valid because random sampling will not change the intrinsic character of sets.

We remark that this subspace loss requires no prior knowledge of the number of motions in a scene, so it can be used to train a flow network on a {\it motion-agnostic} dataset. In a single-motion case, it works similarly to the low-rank loss since the optimal loss is closely related to the rank of ${\bf H}$~\cite{ji2014efficient}. In a multi-motion case, as long as the epipolar subspaces are disjoint and principle angles between them are below certain thresholds~\cite{elhamifar2010clustering}, this loss can still serve as a global regularizer. Even when the scene is highly non-rigid or dynamic, unlike the hard epipolar constraint, this loss won't be detrimental to the network training because it will have same values for both ground-truth flows and wrong flows. In Fig.~\ref{fig:examC}, we show the results of a typical image pair from KITTI using this constraint, demonstrating the effectiveness of our method.

\section{Unsupervised Learning of Optical Flow}
We formulate our unsupervised optical flow estimation approach as an optimization of image based losses and epipolar constraint losses. In unsupervised optical flow estimation, only photometric loss $\mathcal{L}_{\text{photo}}$ can provide data term. Additionally, we use a smoothness term $\mathcal{L}_{\text{smooth}}$ and our epipolar constraint term $\mathcal{L}_{\text{\textbf{F}}|\text{lowrank}|\text{subspace}}$ as our regularization terms. Our overall loss $\mathcal{L}$ is a linear combination of these three losses
\begin{equation}
    \mathcal{L} = \mathcal{L}_{\text{photo}} + \mu_1\mathcal{L}_{\text{smooth}} + \mu_2\mathcal{L}_{\text{\textbf{F}}|\text{lowrank}|\text{subspace}},
    \label{eq:all_function}
\end{equation}
where $\mu_1,\mu_2$ are the weights for each term. We empirically set $\mu_1 = 0.02$ and $\mu_2 = {0.02,0.01,0.001}$ for $\mathcal{L}_{\text{\bf F}}, \mathcal{L}_{\text{lowrank}}^*, \mathcal{L}_{\text{subspace}}$ respectively.

\subsection{Image Warping Loss}
Similarly to conventional methods, we leverage the most popular brightness constancy assumption, \ie, $I^t, I^{t+1}$ should have similar pixel intensities, colors and gradients. Our photometric error is then defined by the difference between the reference frame and the warped target frame based on flow estimation. 

In \cite{Simon2018}, they target at the case which the illumination may changes from frame to frame and propose a bidirectional census transform $C(\cdot)$ to handle this situation. We adopt this idea to our photometric error. Therefore, our photometric loss is a weighted summation of pixel intensities (or color) loss $\mathcal{L}_{i}$, image gradient loss $\mathcal{L}_{g}$ and bidirectional census loss $\mathcal{L}_{c}$.
\begin{equation}
\mathcal{L}_{\text{photo}} = \lambda_{1}\mathcal{L}_{i}+\lambda_{2}\mathcal{L}_{c}+ \lambda_{3}\mathcal{L}_{g},
\end{equation}
where $\lambda_1 = 0.5,\lambda_2 = 1,\lambda_{3} = 1$ are the weights for each term.

Inspired by \cite{Wang_2018_CVPR}, we only compute our photometric loss on non-occluded areas $O$ and normalize the loss by the number of pixels of non-occluded regions. We determine a pixel to be occluded or not by forward-backward consistency check. If the sum of its forward and backward flow is above a threshold $\tau$, we set the pixel as occluded. We use $\tau = 3$ in all experiments.

Our photometric loss is thus defined as follows:
\begin{equation}
\mathcal{L}_i = \left[\sum_{i=1}^NO_i\cdot\varphi(\hat{I}^t(\mathbf{x}_i)-I^t(\mathbf{x}_i))\right]/\sum_i^NO_i
\end{equation}
\begin{equation}
\mathcal{L}_c = \left[\sum_{i=1}^NO_i\cdot\varphi(\widehat{C}^t(\mathbf{x}_i)-C^t(\mathbf{x}_i))\right]/\sum_i^NO_i
\end{equation}
\begin{equation}
\mathcal{L}_g = \left[\sum_{i=1}^NO_i\cdot\varphi(\nabla \hat{I}^t(\mathbf{x}_i)-\nabla I^t(\mathbf{x}_i))\right]/\sum_i^NO_i
\end{equation}
where $\hat{I}^t(\mathbf{x}_i) = I^{t+1}(\mathbf{x}_i+\mathbf{v}_i)$ is computed through image warping with the estimated flow, and following \cite{Wang_2018_CVPR}, we use a robust Charbonnier penalty $\varphi(x) = \sqrt{x^{2} + 0.001^{2}}$ to evaluate differences.  

\subsection{Smoothness Loss}
Commonly, there are two kinds of smoothness prior in conventional optical flow estimation: One is piece-wise planar, and the other is piece-wise linear. The first one can be implemented by penalizing the first order derivative of recovered optical flow and the later one is by the second order derivative. For most rigid scenes, piece-wise planar model can provide a better interpolation. But for deformable cases, piece-wise linear model suits better. Therefore, we use a combination of these two models as our smoothness regularization term. We further assumes that the edges in optical flows are edges in reference color images as well. Formally, our image guided smoothness term can be defined as:  
\begin{equation}
\mathcal{L}_{s} = \sum\left(e^{-\alpha_1\left|\nabla I\right|}\left|\nabla V\right|+e^{-\alpha_2\left|\nabla^2 I\right|}\left|\nabla^2 V\right|\right)/N,
\end{equation}
where $\alpha_1 = 0.5$ and $\alpha_2 = 0.5$ and $V\in \mathbb{R}^{W\times H\times2}$ is a matrix form of $\mathbf{v}$.

\section{Experiments}
We evaluate our methods on standard optical flow benchmarks, including KITTI \cite{Geiger2012CVPR,Menze2015CVPR}, MPI-Sintel \cite{Butler:ECCV:2012}, Flying Chairs \cite{Dosovitskiy:2015}, and Middlebury \cite{Baker2011}. We compare our results with existing optical flow estimation methods based on standard metrics, \ie, endpoint error (EPE) and percentage of optical flow outliers (Fl). We denote our method as EPIFlow.

\subsection{Implementation details.}
\paragraph{Architecture and Parameters.} 
We implemented our EPIFlow network in an end-to-end manner by adopting the architecture of PWC-Net \cite{Sun2018PWC-Net} as our base network due to its state-of-the-art performance. The original PWC-Net takes a structure of pyramid and learns on 5 different scales. However, a warping error is ineffective on low resolutions. Therefore, we pick the highest resolution output, upsample it to match the input resolution by bilinear interpolation, and compute our self-supervised learning losses only on that scale. The learning rate for initial training (from scratch) is $10^{-4}$  and that for fine-tuning is $10^{-5}$. Depending on the resolution of input images, the batch size is 4 or 8. We use the same data argumentation scheme as proposed in FlowNet2~\cite{Ilg17}.
Our network's typical speed varies from 0.07 to 0.25 seconds per frame during the training process, depending on the input image size and the losses used, and is around 0.04 seconds per frame in evaluation. The experiments were tested on a regular computer equipped with a Titan XP GPU. EPIFlow is significantly faster compared with conventional methods.
\begin{table*}[t]
\setlength{\tabcolsep}{1.5mm}
	\begin{tabular}{l l c c c c c c c c c c c c}
		\cmidrule(lr){1-13}
		 & & \multicolumn{4}{c}{KITTI 2012} & \multicolumn{3}{c}{KITTI 2015} & \multicolumn{2}{c}{Sintel Clean}&\multicolumn{2}{c}{Sintel Final}\\
        \cmidrule(lr){3-6}
        \cmidrule(lr){7-9}
        \cmidrule(lr){10-11}
        \cmidrule(lr){12-13}
        &Method & \multicolumn{2}{c}{EPE(all)}&\multicolumn{2}{c}{EPE(noc)}& EPE(all)& EPE(noc)&Fl$-$all & \multicolumn{2}{c}{EPE(all)}&\multicolumn{2}{c}{EPE(all)}\\
        \cmidrule(lr){3-4}
        \cmidrule(lr){5-6}
        \cmidrule(lr){7-7}
        \cmidrule(lr){8-8}
        \cmidrule(lr){9-9}
        \cmidrule(lr){10-11}
        \cmidrule(lr){12-13}
        & & \textit{train} & \textit{test} &  \textit{train} & \textit{test} & \textit{train} & \textit{train} & \textit{test} & \textit{train} & \textit{test} & \textit{train} & \textit{test}\\
		\cmidrule(lr){1-13}
        \parbox[t]{2mm}{\multirow{2}{*}{\rotatebox[origin=c]{90}{\tiny{Non-deep}}}}&EpicFlow \cite{DBLP:journals/corr/RevaudWHS15} & 3.47 & 3.8 & -- & 1.5 & 9.27 &--& 26.29\% & 2.27& 4.11& 3.56&6.29\\
        &MRFlow \cite{Wulff:CVPR:2017} & --& --& --& --& --& --&12.19\%&(1.83)&2.53&(3.59)&5.38 \\
        \cmidrule(lr){1-13}
        \parbox[t]{2mm}{\multirow{4}{*}{\rotatebox[origin=c]{90}{\small{Supervised}}}}
&SpyNet-ft \cite{DBLP:journals/corr/RanjanB16}&(4.13) &4.1 &-- & 2.0 &-- &-- &35.07\%&(3.17)&6.64 &(4.32)&8.36\\
        &FlowNet2-ft \cite{Ilg17} & (1.28) & 1.8 & --& 1.0 & 2.30 &--& 10.41\%&(1.45) &4.16&(2.01) &5.74  \\
        &PWC-Net \cite{Sun2018PWC-Net} & 4.14 & -- & --& -- & 10.35& -- &--  &2.55 & -- &3.93 & --\\             &PWC-Net-ft \cite{Sun2018PWC-Net} & (1.45) & 1.7 & --& 0.9 & (2.16)& -- &9.60\%  &(1.70) & 3.86 &(2.21) &5.17\\ 
        \cmidrule(lr){1-13}
        \parbox[t]{2mm}{\multirow{13}{*}{\rotatebox[origin=c]{90}{Unsupervised}}}&UnsupFlownet \cite{Yu2016} & (11.30) &9.9& (4.30)& 4.6& -- & -- &-- & --&  -- &-- &--\\
        &DSTFlow-ft \cite{ren2017unsupervised} & (10.43) &12.4 &(3.29) &4.0 &(16.79) & (6.96) &39.00\%  &(6.16)&10.41&(7.38)&11.28\\
        &DF-Net-ft \cite{zou2018df}&(3.54)&4.4&-- &--&(8.98)&--&25.70\%& --&--&--&--\\
        &GeoNet \cite{yin2018geonet} & --&--& --&--&10.81&8.05&--&--&--&--&--\\ 
        &UnFlow \cite{Simon2018} & (3.29) & --& (1.26)&-- & (8.10)&--&--& --&  9.38 &7.91& 10.21\\
        &OAFlow-ft \cite{Wang_2018_CVPR}& (3.55) & 4.2&--&-- & (8.88)& -- & 31.20\%&(4.03)  & 7.95 &(5.95)& 9.15 \\  
     
        &CCFlow \cite{ranjan2018adversarial} &--&--&--&--&(5.66)&--&25.27\%&--&--&--&-- \\        
        &Back2Future-ft \cite{Janai2018ECCV} & --& --& --& --&(6.59)&(3.22)&22.94\%&(3.89)&7.23&(5.52) &8.81\\
        \cmidrule(lr){2-13}
        &Our-baseline & 3.23 &--& 1.04&-- & 7.93& 4.21& --&6.72 & --&7.31& --\\  
        &Our-gtF & 2.61 & --& 1.04 & --& 6.03& 2.89& --&6.15 & --&6.71& --\\
        &Our-F & 2.56 & --& \textbf{0.97 }& --& 6.42& 3.09& --&6.21 & --&6.73& --\\
        &Our-low-rank & 2.63 & --& 1.07 & --& 5.91& 3.03& --&6.39 & --&6.96& --\\       
        &Our-sub & 2.62 & --& 1.03 & --& 6.02& 2.98 &--&6.15 & --&6.83& --\\
        &Our-sub-test-ft & 2.61&(\textbf{3.2})&1.03&(\textbf{1.1})&5.56 &2.56  &(\textbf{16.24\%}) & 3.94& (\textbf{6.84})&5.08&(\textbf{8.33})\\  
        &Our-sub-train-ft & (\textbf{2.51})&3.4&(0.99)&1.3&(\textbf{5.55}) &(\textbf{2.46})  &16.95\% & \textbf{(3.54)}& 7.00&\textbf{(4.99)}&8.51\\  
		\cmidrule(lr){1-13}
	\end{tabular}
    \caption{\textbf{Performance comparison on the KITTI and Sintel optical flow benchmarks.} The metric EPE(noc) indicates the average endpoint error of non-occluded regions while the term EPE(all) is that for all pixels. The KITTI 2015 testing dataset evaluates results by the percentage of flow outliers (Fl). The baseline, gtF, F, low-rank, and sub models were trained on the KITTI VO dataset. 
    The parentheses indicate the corresponding models that were trained on the same data and the missing entries (-) indicate the results were not reported.  
    Note that the current STOA unsupervised method Back2Future Flow \cite{Janai2018ECCV} uses three frames as input. Best results are marked by bold fonts.}
\label{tab:results}
\vspace{-0.2cm}
\end{table*}

\paragraph{Pre-training.}
We pre-trained our network on the Flying Chairs dataset using a weighted combination of the warping loss and smoothness loss. Flying Chairs is a synthetic dataset consisting of rendered chairs superimposed on real-world Flickr images. Training on such a large-scale synthetic dataset allows the network to learn the general concepts of optical flow before handling complicated real-world conditions, \eg, changeable light or motions. To avoid trivial solutions, we disabled the occlusion-aware term at the beginning of the training (\ie, the first two epochs). Otherwise, the network would generate all zero occlusion masks which invalidate losses. The pre-training roughly took forty hours and its returned model was used as an initial model for other datasets. 

\subsection{Datasets}
\paragraph{KITTI Visual Odometry (VO) Dataset.} 
The KITTI VO dataset contains 22 calibrated sequences with 87,060 consecutive pairs of real-world images. The ground truth poses of the first 11 sequences are available. We fine-tuned our initial model on the KITTI VO dataset using various loss combinations. We chose it for two reasons: (1) it provides ground truth camera poses for every frame, which simplifies the problem of network performance analysis and (2) most scenes in the KITTI VO dataset are stationary and thus can be fitted by an ego-motion. The relative poses (between a pair of images) and camera calibration can be used to compute fundamental matrices. To compare our various methods fairly, we use the first 11 sequences as our training set. 

\vspace{-4mm}
\paragraph{KITTI Optical Flow Dataset.} The KITTI optical flow dataset contains two subsets: KITTI 2012 and KITTI 2015, where the first one mostly contains stationary scenes and the latter one includes more dynamic scenes. KITTI 2012 provides 194 annotated image pairs for training and 195 pairs for testing while KITTI 2015 provides 200 pairs for training and 200 pairs for testing. Our training did not use the KITTI datasets' multiple-view extension images.

\vspace{-4mm}
\paragraph{MPI Sintel Dataset.} The MPI Sintel dataset provides naturalistic frames which were captured from an open source movie. It contains 1041 training image pairs with ground truth optical flows and pixel-wise occlusion masks, and also provides 552 image pairs for benchmark testing. The scenes of the MPI Sintel dataset were rendered under two different complexity (Clean and Final). Unlike the KITTI datasets, most scenes in the Sintel dataset are highly dynamic.
\begin{figure*}[t]
    \centering
    \tabcolsep=0.05cm
    \begin{tabular}{c c c c c}
Input & Ours & Back2Future \cite{Janai2018ECCV} &Our Error& Back2Future Error \cite{Janai2018ECCV}\\
    \includegraphics[width=3.4cm]{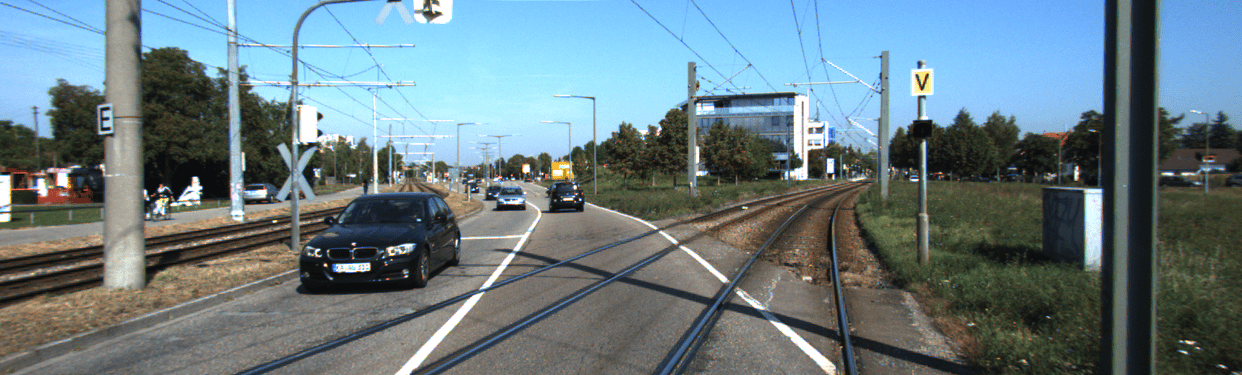}  &
    \includegraphics[width=3.4cm]{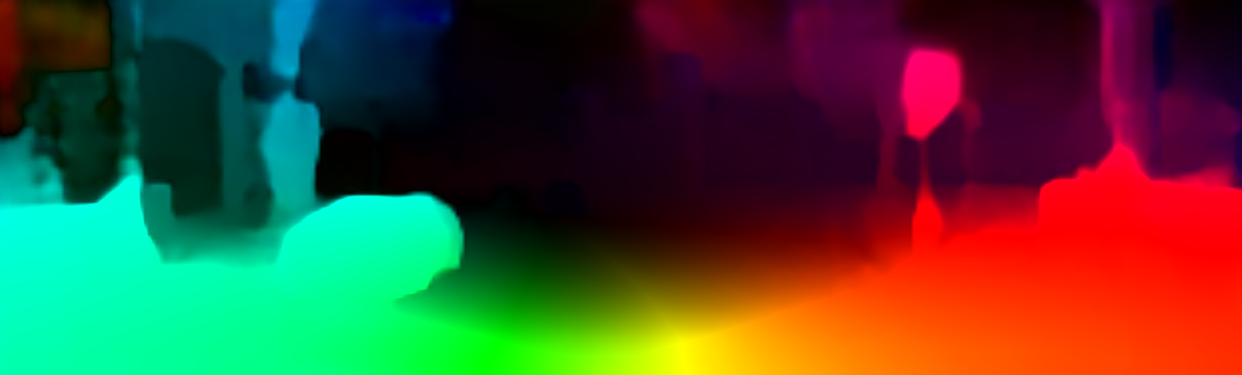} &
        \includegraphics[width=3.4cm]{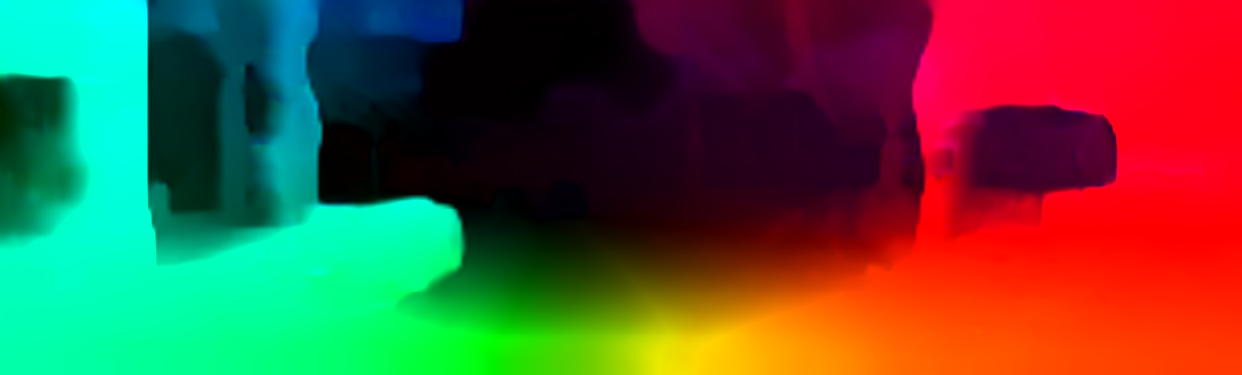} &
\includegraphics[width=3.4cm]{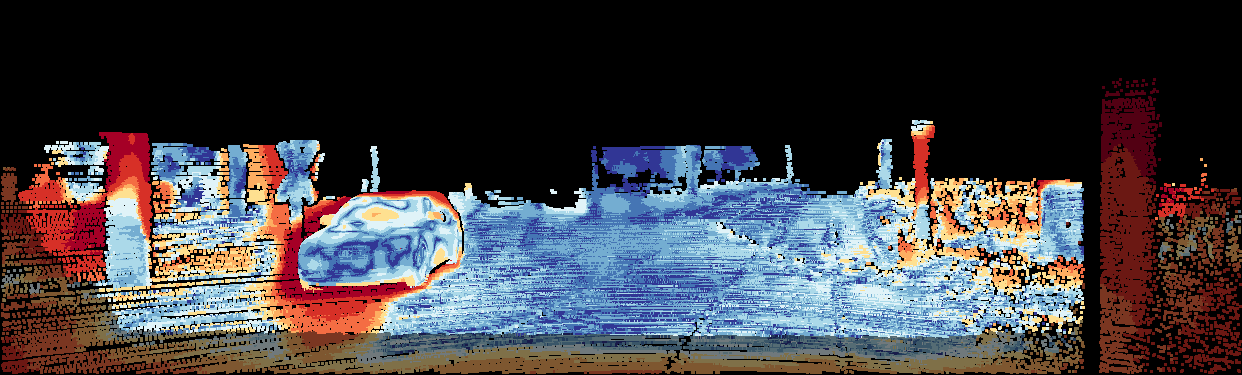} &        \includegraphics[width=3.4cm]{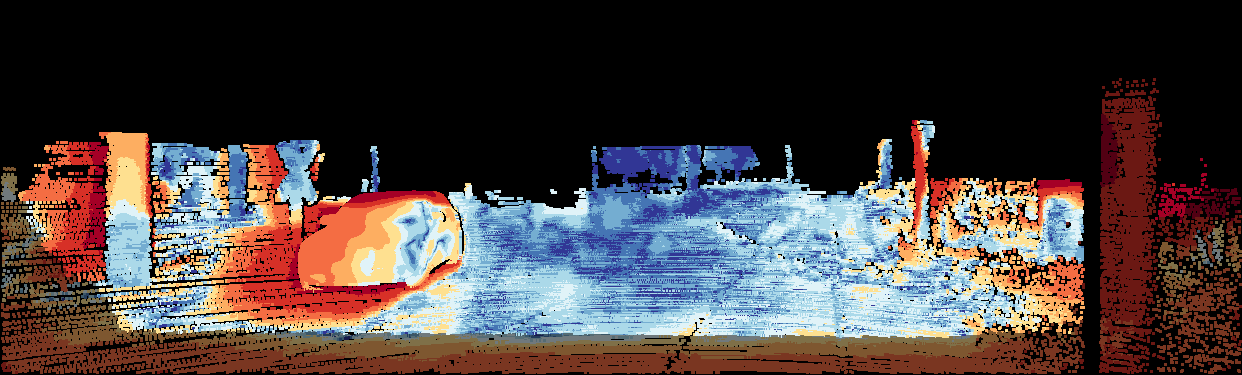} \\
    \includegraphics[width=3.4cm]{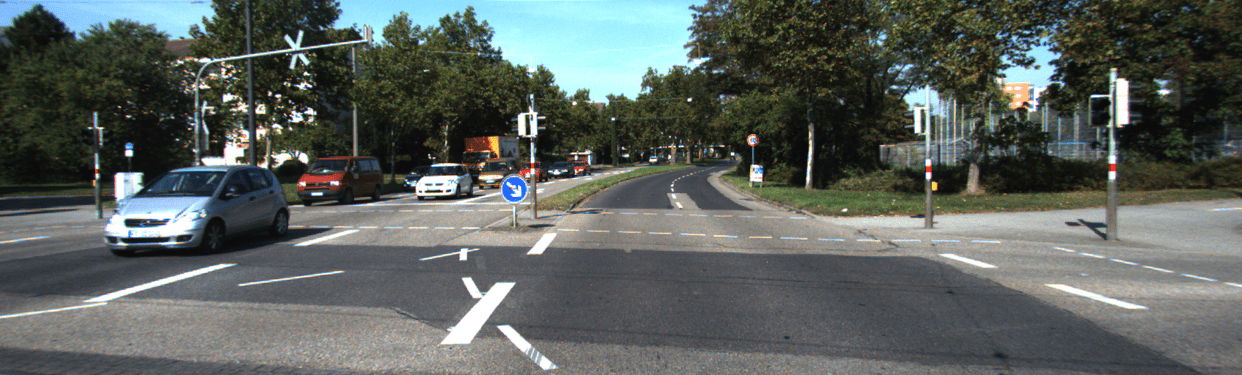}  &
    \includegraphics[width=3.4cm]{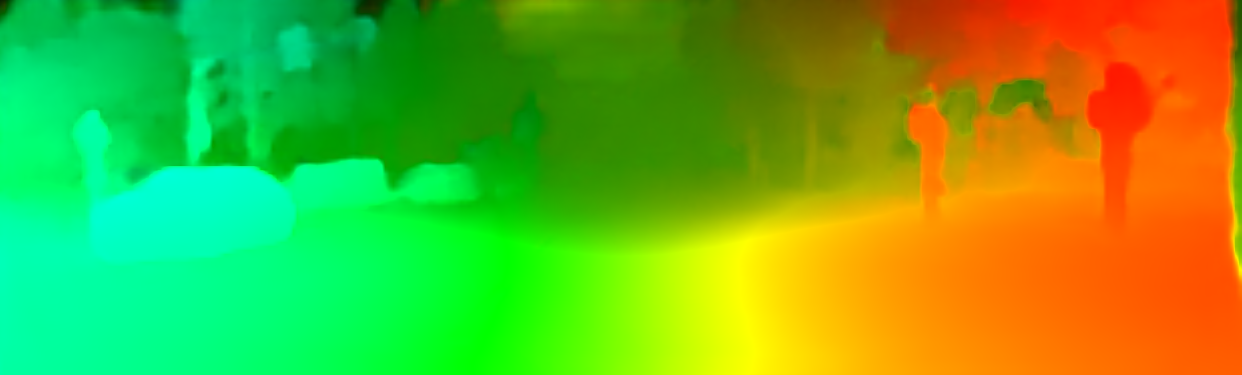} &
        \includegraphics[width=3.4cm]{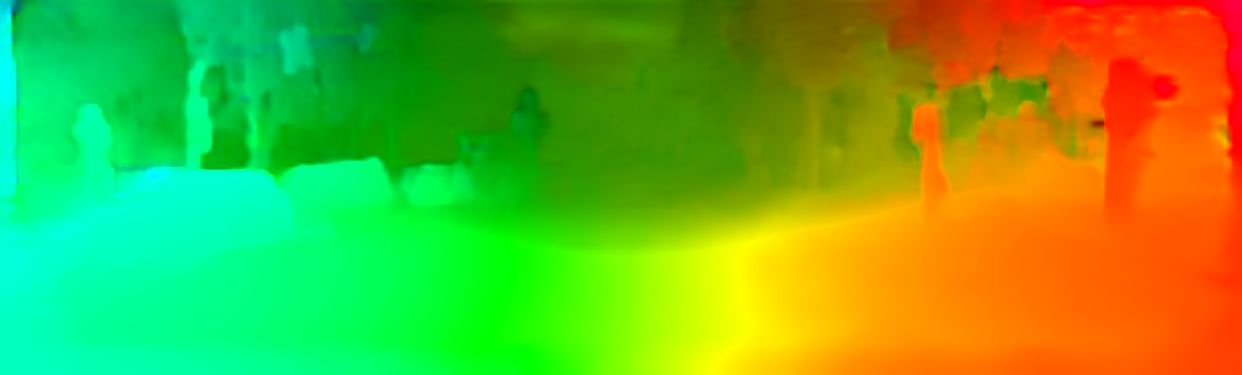} &
\includegraphics[width=3.4cm]{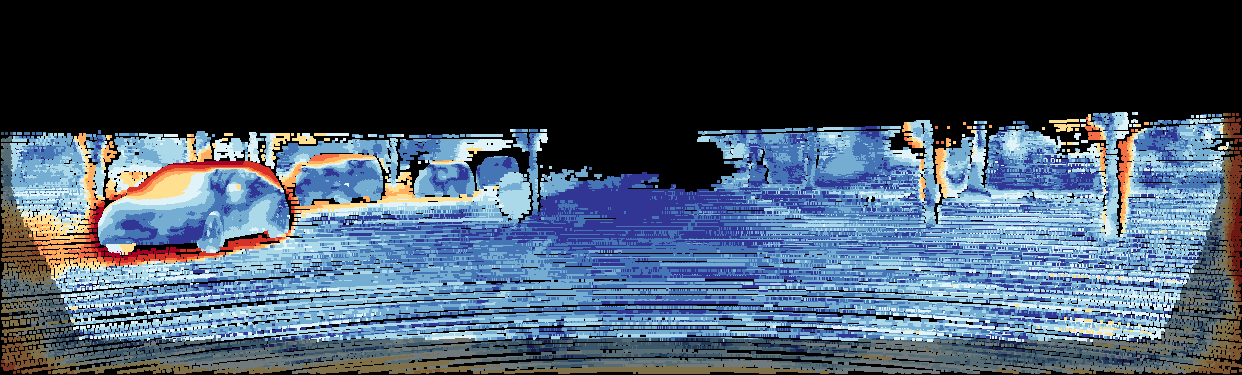} &        \includegraphics[width=3.4cm]{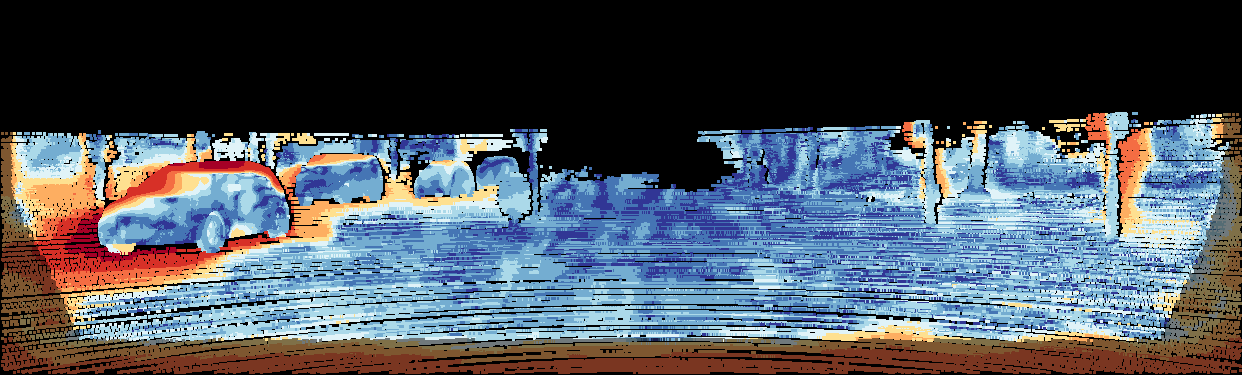} \\
    \includegraphics[width=3.4cm]{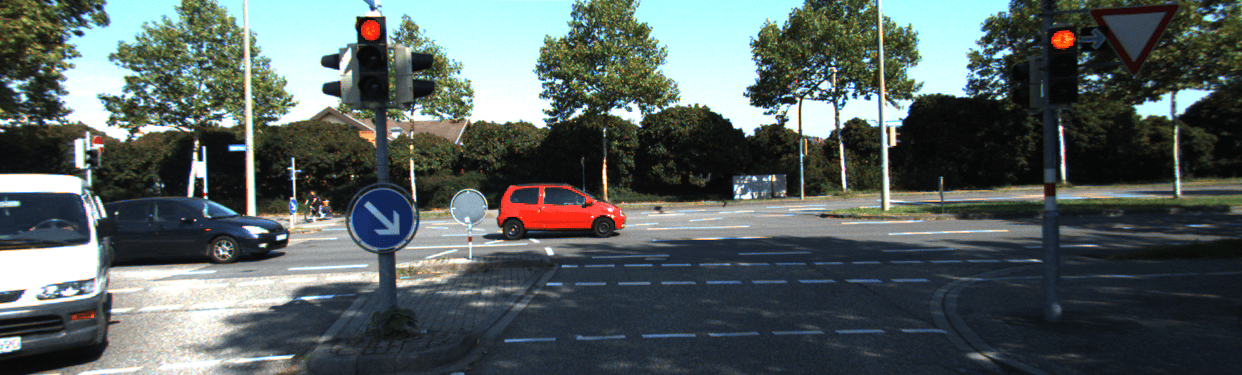}  &
    \includegraphics[width=3.4cm]{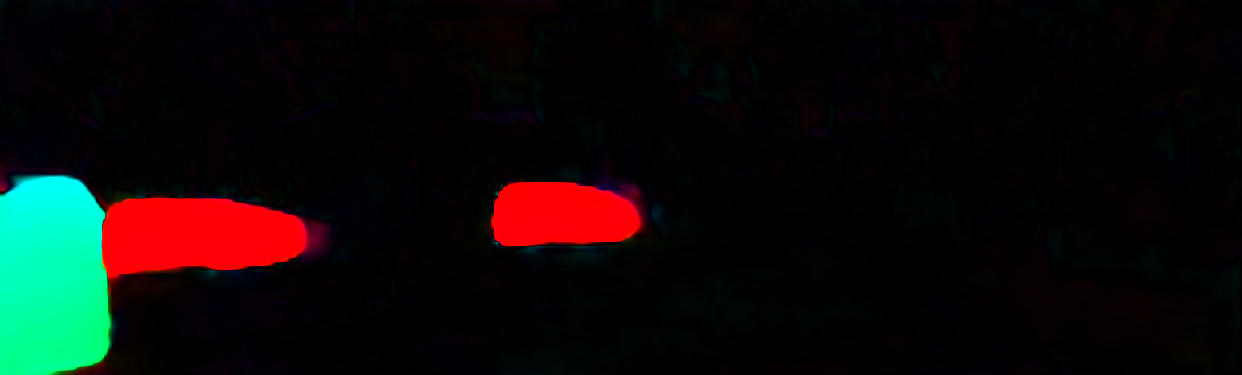} &
        \includegraphics[width=3.4cm]{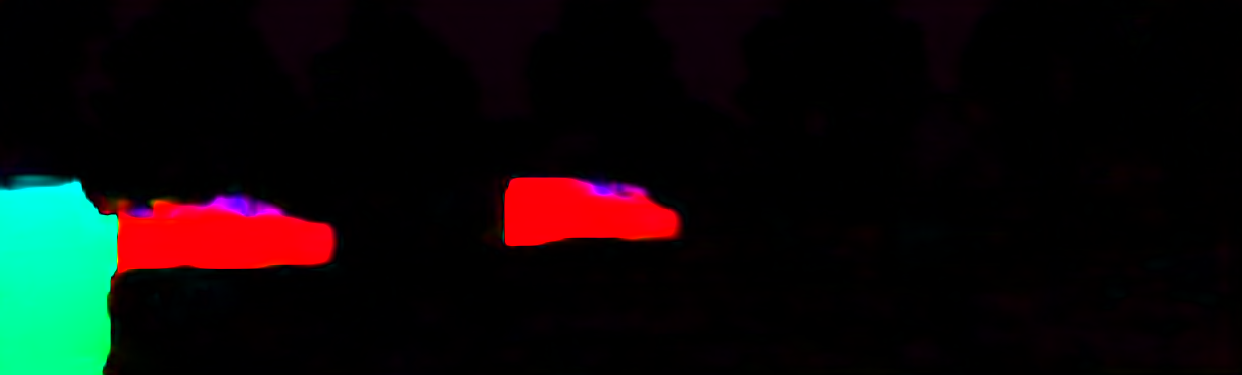} &
\includegraphics[width=3.4cm]{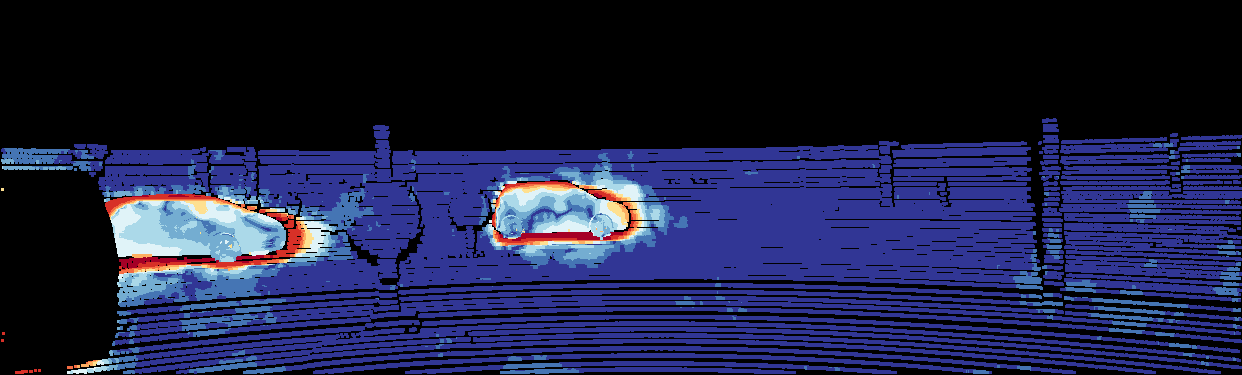} &        \includegraphics[width=3.4cm]{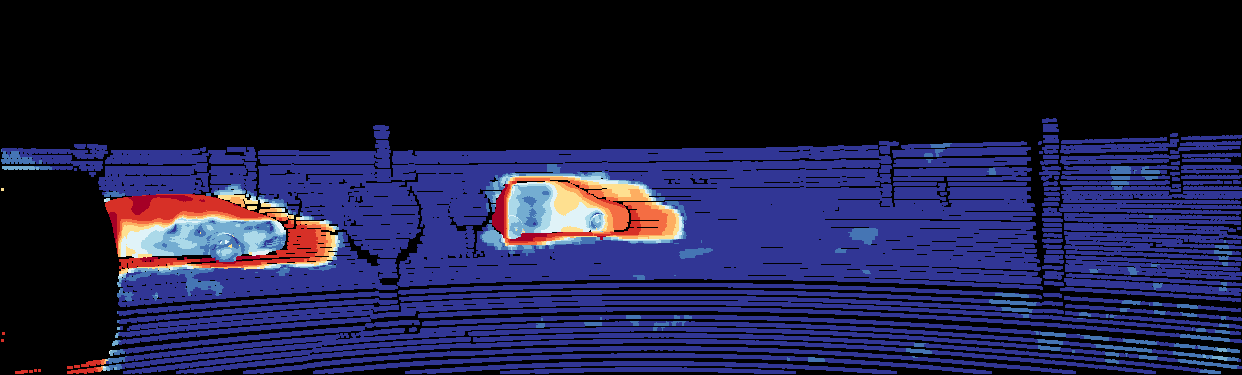} \\
    \includegraphics[width=3.4cm]{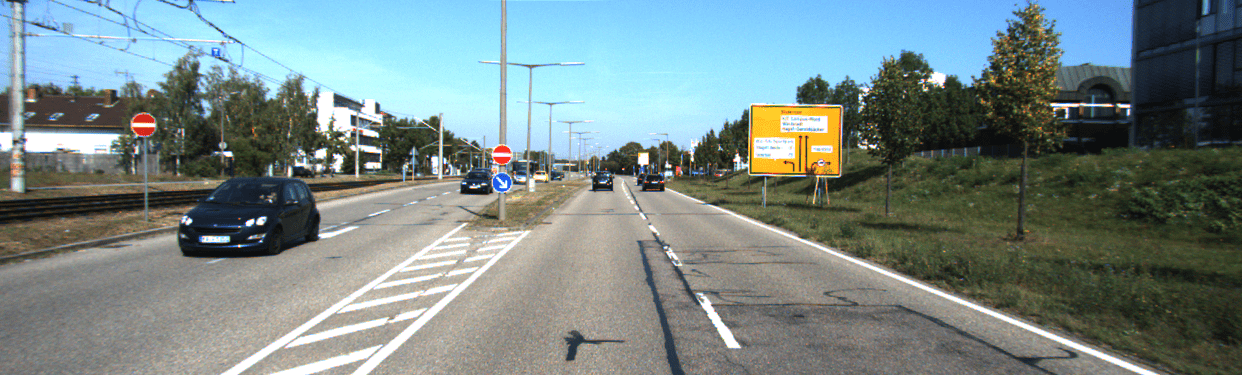}  &
    \includegraphics[width=3.4cm]{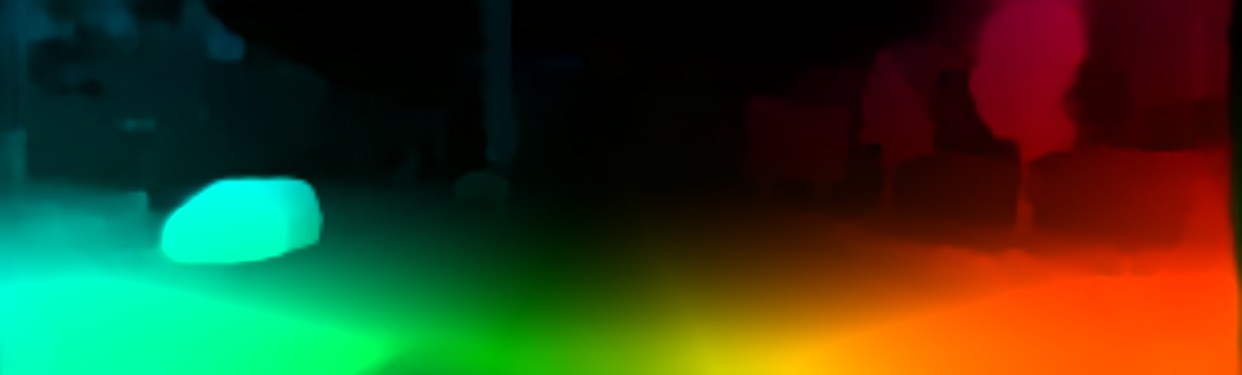} &
        \includegraphics[width=3.4cm]{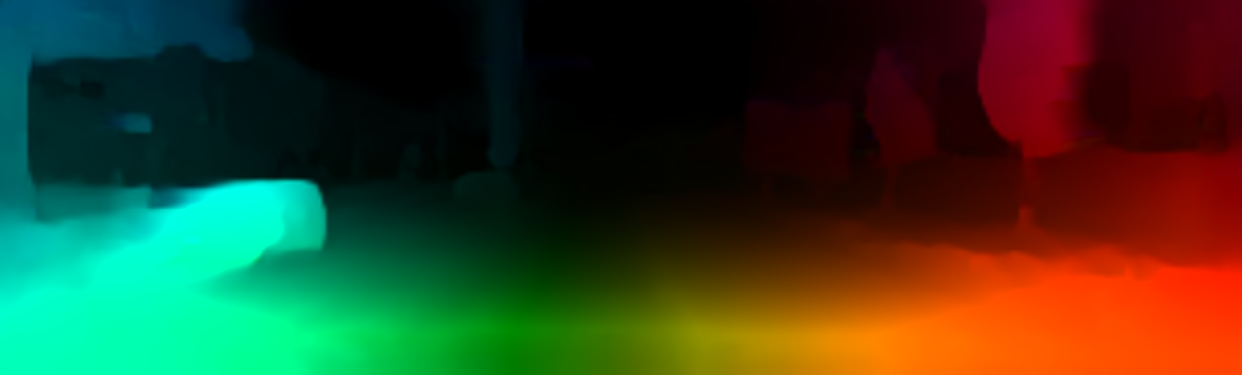} &
\includegraphics[width=3.4cm]{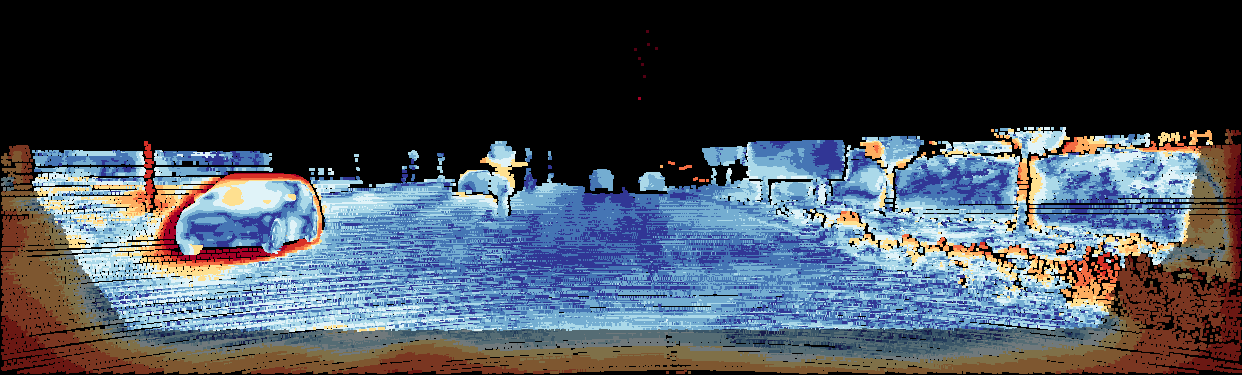} &        \includegraphics[width=3.4cm]{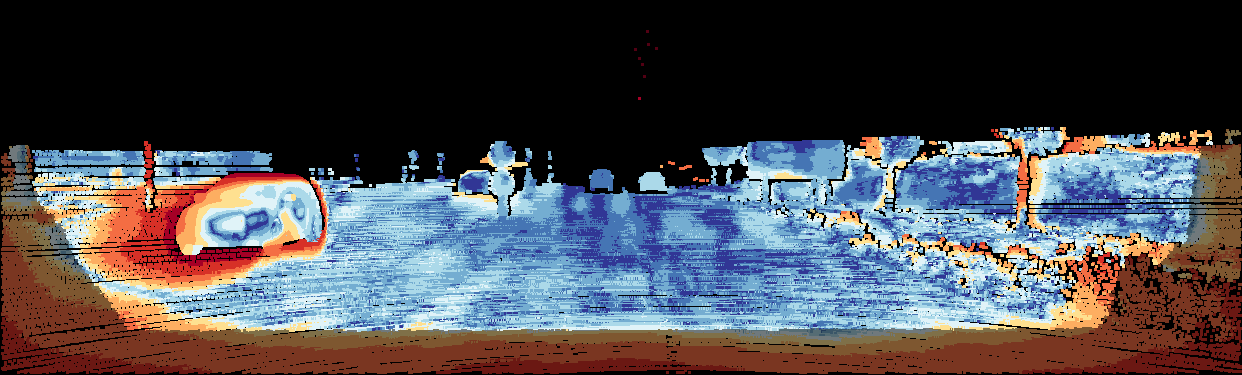} \\
    \end{tabular}
	\caption{\textbf{Qualitative results on KITTI 2015 Test dataset.} We compare our method with Back2Future Flow \cite{Janai2018ECCV}. The second column contains the flows estimated by Our-sub-ft model while the third column contains the results of Back2Future Flow. The flow error visualization is also provided where correct estimates are depicted in blue and wrong ones in red. Consistent with the quantitative analysis, our results are visually better on structural boundaries }
    \label{fig:kitti_results}
    \vspace{-0.3cm}
\end{figure*}

\begin{figure*}
    \centering
    \tabcolsep=0.05cm
    \begin{tabular}{c c c c c}
Input & Ours & Back2Future \cite{Janai2018ECCV}&Our Error& Back2Future Error \cite{Janai2018ECCV}\\
    \includegraphics[width=3.4cm]{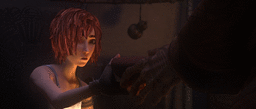}  &
    \includegraphics[width=3.4cm]{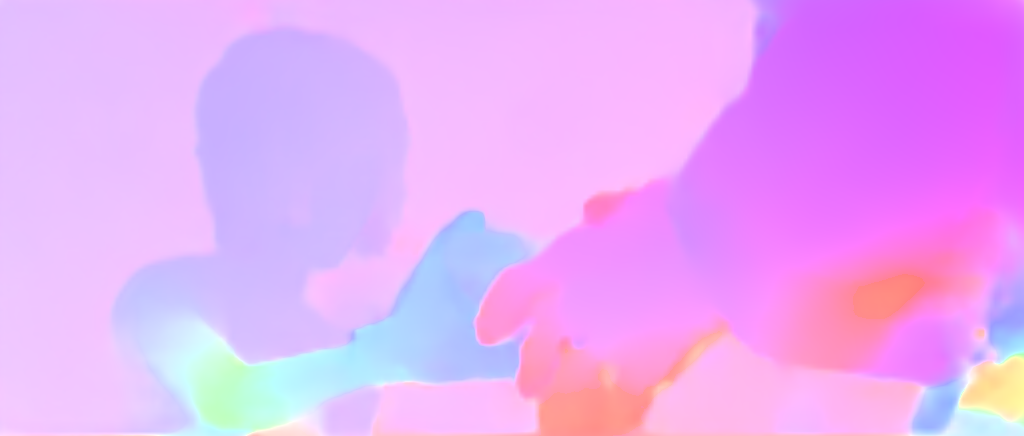} &
        \includegraphics[width=3.4cm]{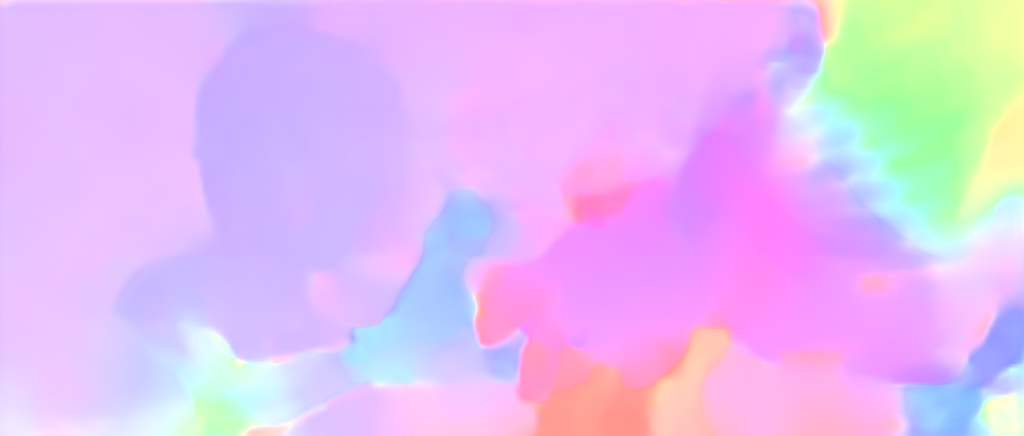} &
\includegraphics[width=3.4cm]{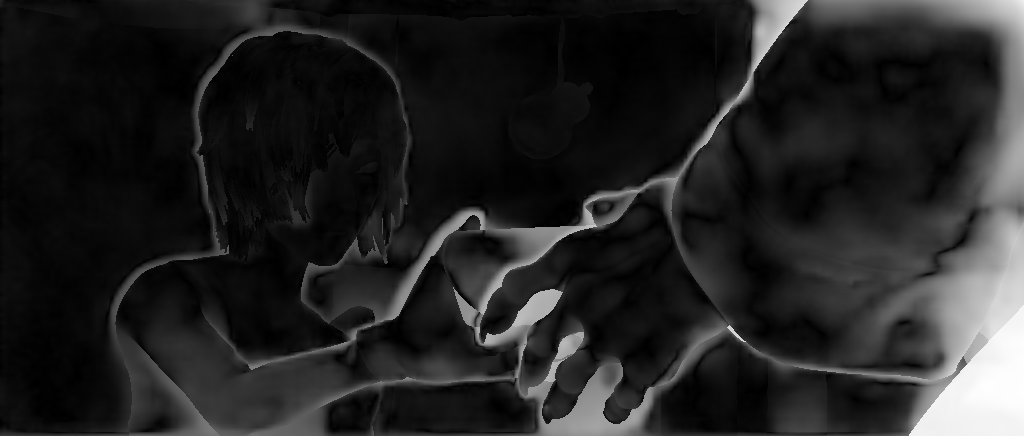} &        \includegraphics[width=3.4cm]{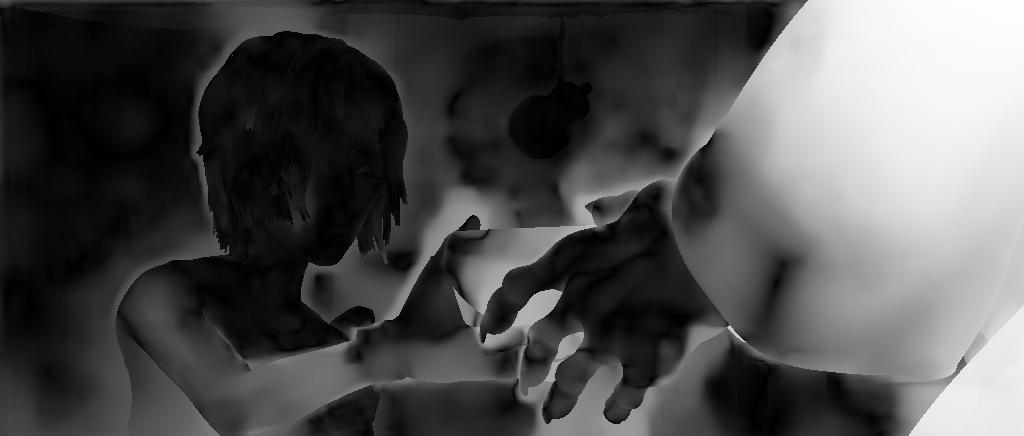} \\
    \includegraphics[width=3.4cm]{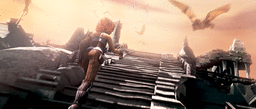}  &
    \includegraphics[width=3.4cm]{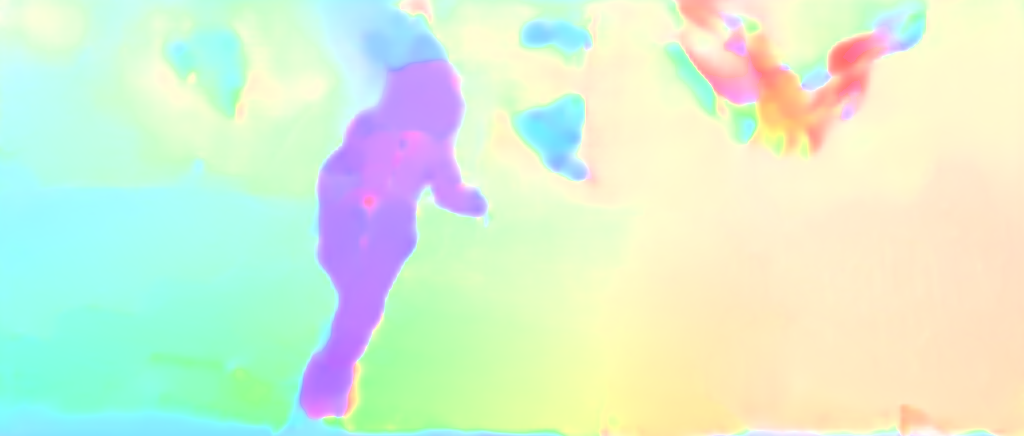} &
        \includegraphics[width=3.4cm]{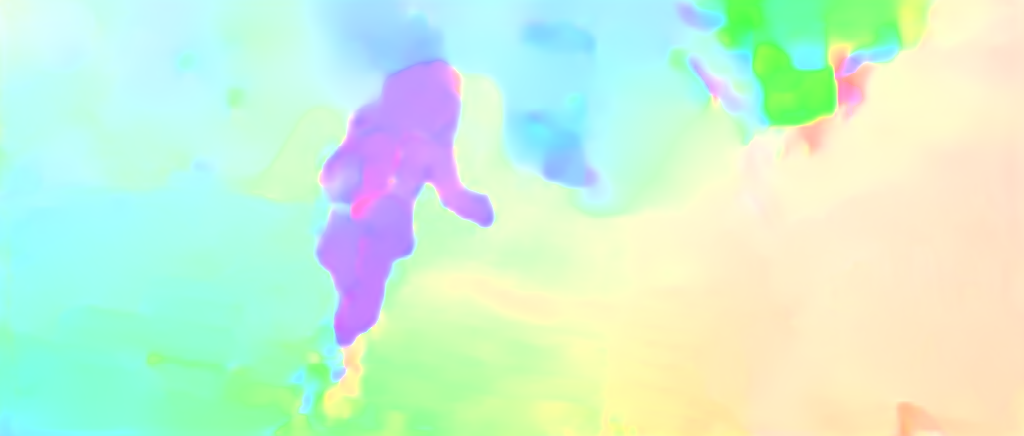} &
\includegraphics[width=3.4cm]{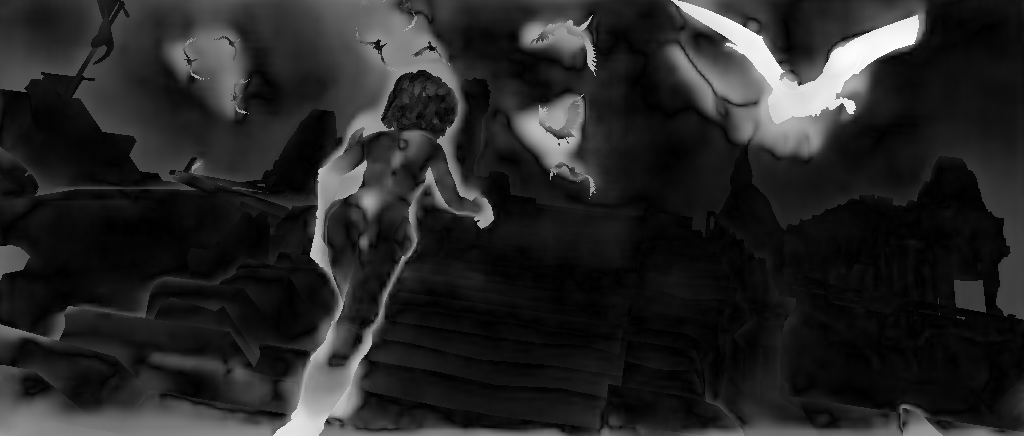} &        \includegraphics[width=3.4cm]{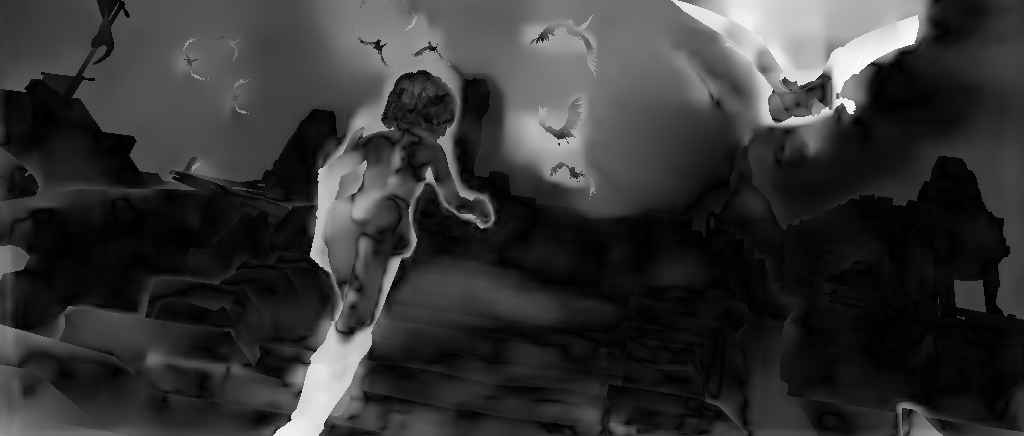} \\
    \includegraphics[width=3.4cm]{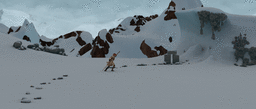}  &
    \includegraphics[width=3.4cm]{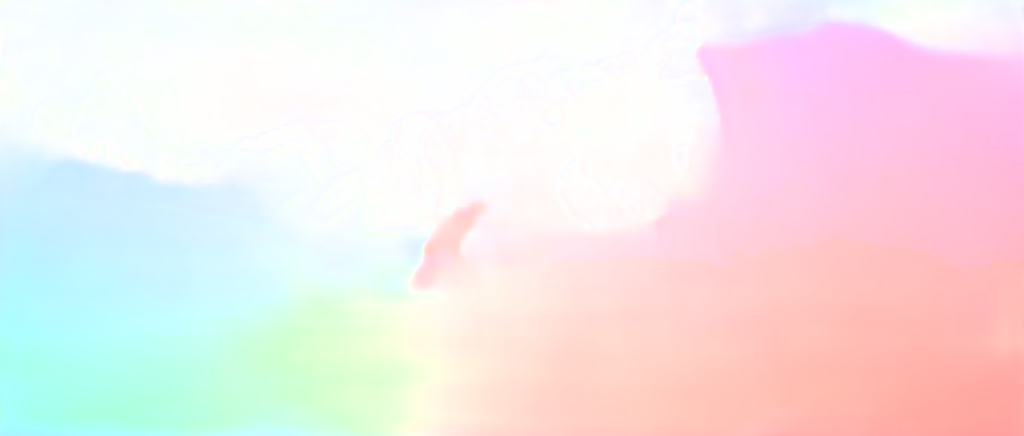} &
        \includegraphics[width=3.4cm]{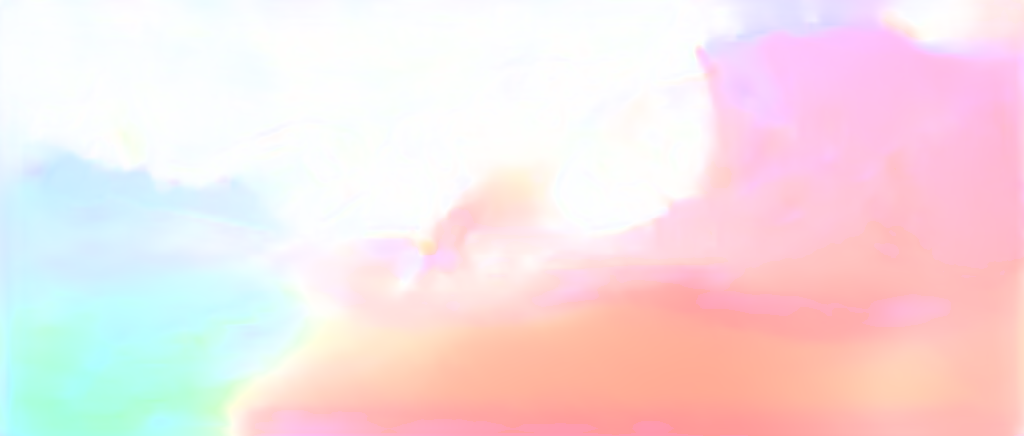} &
\includegraphics[width=3.4cm]{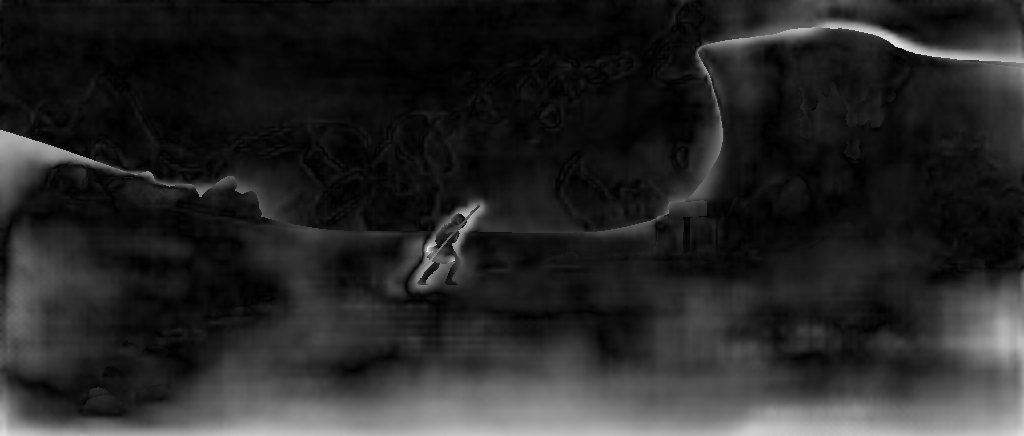} &        \includegraphics[width=3.4cm]{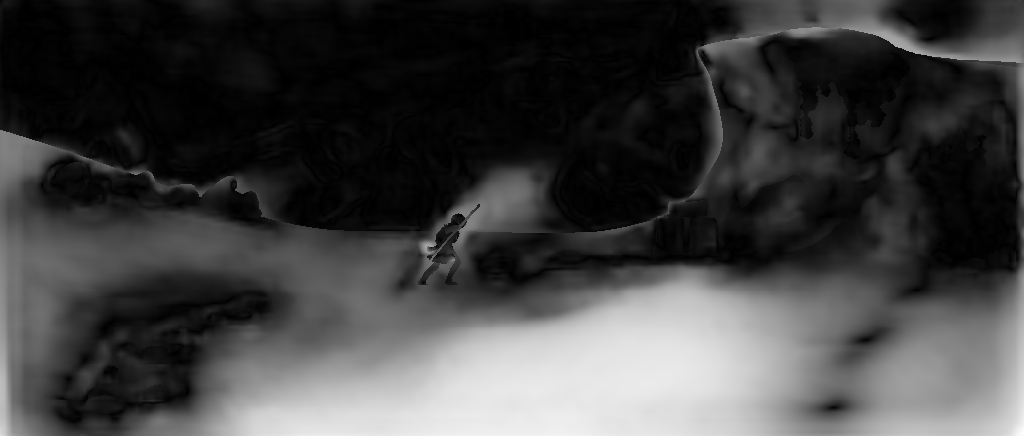} \\
    \includegraphics[width=3.4cm]{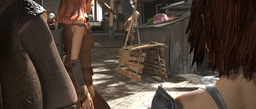}  &
    \includegraphics[width=3.4cm]{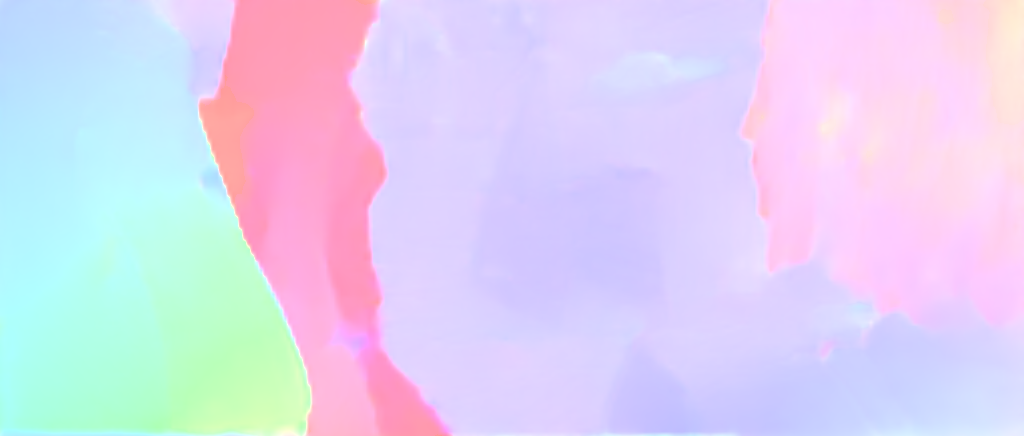} &
        \includegraphics[width=3.4cm]{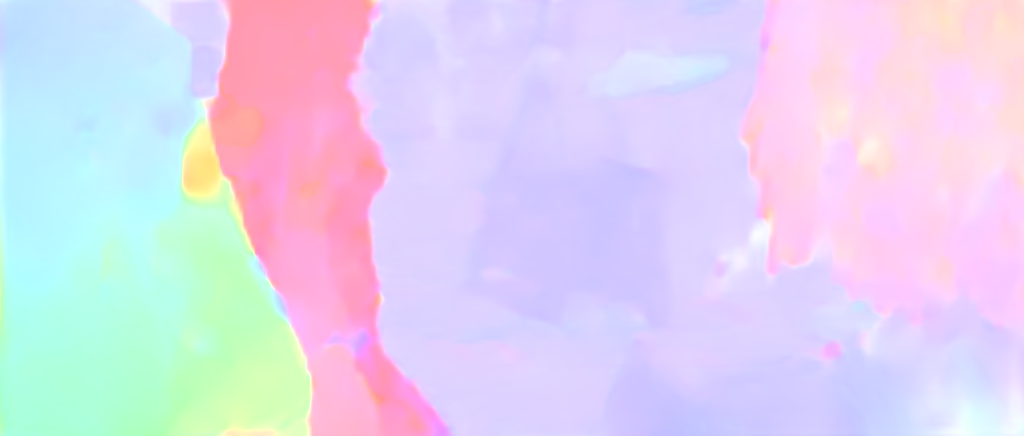} &
\includegraphics[width=3.4cm]{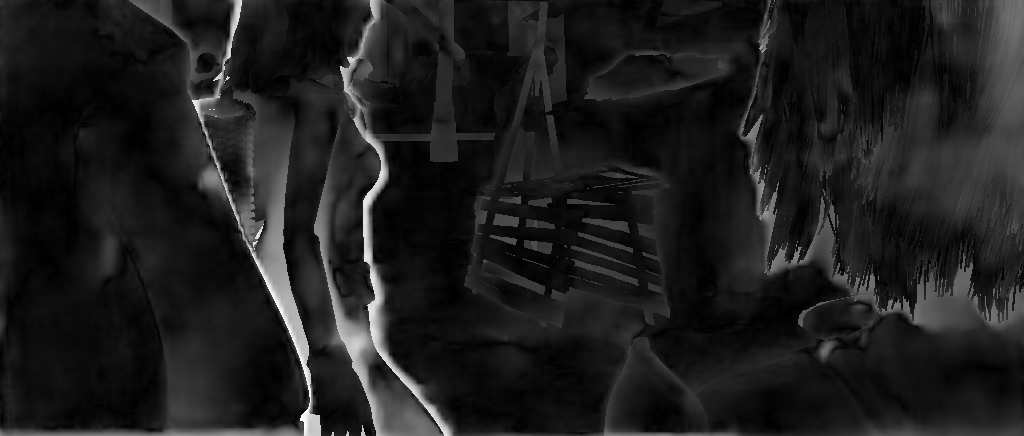} &        \includegraphics[width=3.4cm]{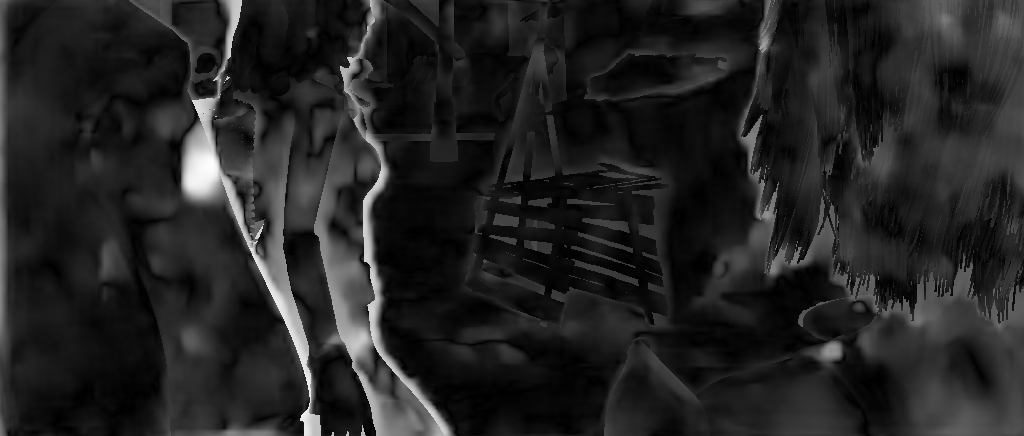} 
    \\

    \end{tabular}
	\caption{\textbf{Qualitative results on the MPI Sintel dataset.} This figure shares the same layout with Fig. \ref{fig:kitti_results} except the top two rows are from the Final set and the two bottom rows are from the Clean set. The errors are visualized in gray on the Sintel benchmark.}
    \label{fig:sintel_results}
    \vspace{-0.3cm}
\end{figure*}

\subsection{Quantitative and Qualitative Results}

We use the suffix ``-baseline'' to indicate our baseline model that was trained using only photometric and smoothness loss. ``-F'' represents the model that was trained using hard fundamental matrix constraint with estimated \textbf{F}. ``-gtF'' means that we used the ground truth fundamental matrix. ``-low-rank'' refers to the model applying the low rank constraint, and ``-sub'' is the model using our subspace constraint. ``-ft'' denotes the model fine-tuned on the datasets. 

\begin{figure*}
    \centering
    \tabcolsep=0.05cm
    \begin{tabular}{c c c c c}
Input & Our-baseline & Our-F&  Our-low-rank &Our-sub\\
    \includegraphics[width=3.4cm]{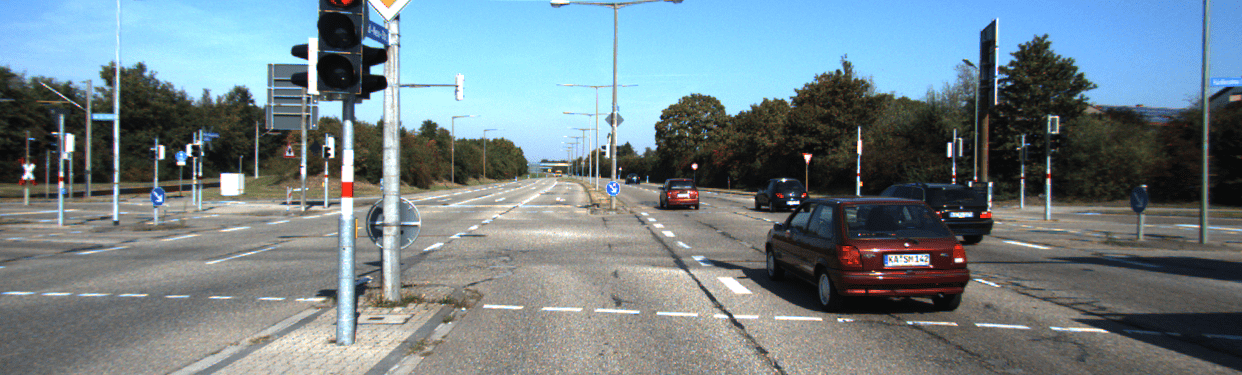}  &
    \includegraphics[width=3.4cm]{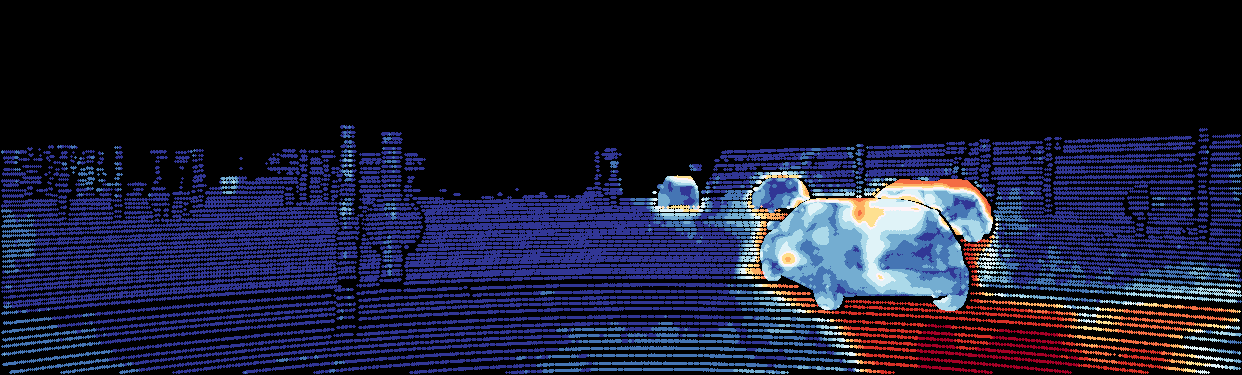} &
        \includegraphics[width=3.4cm]{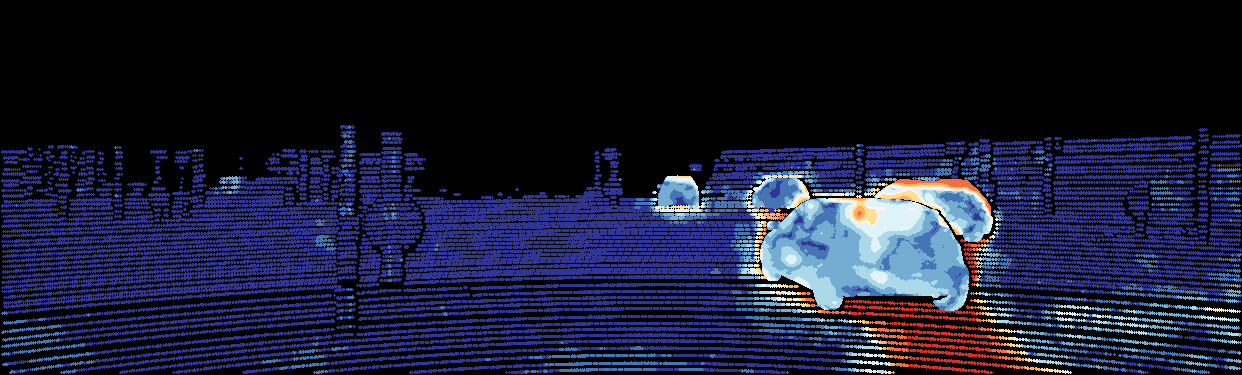} &
    \includegraphics[width=3.4cm]{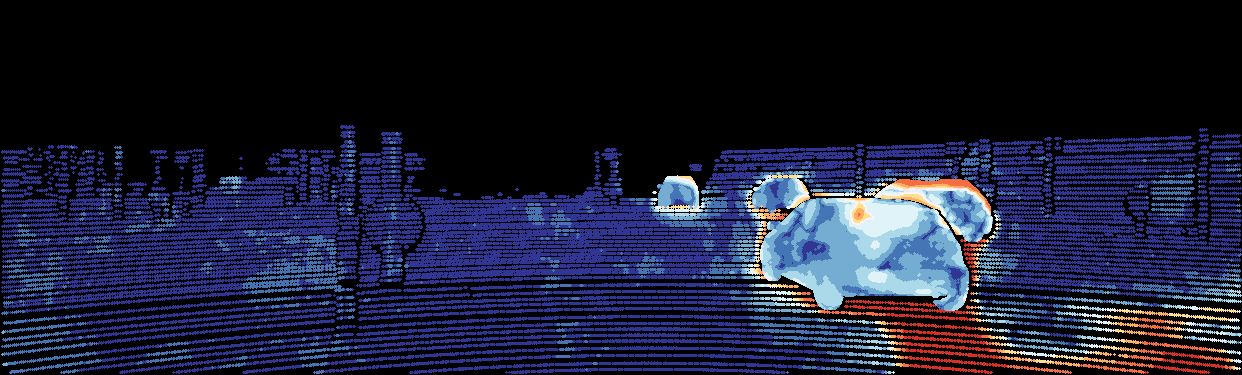} &
    \includegraphics[width=3.4cm]{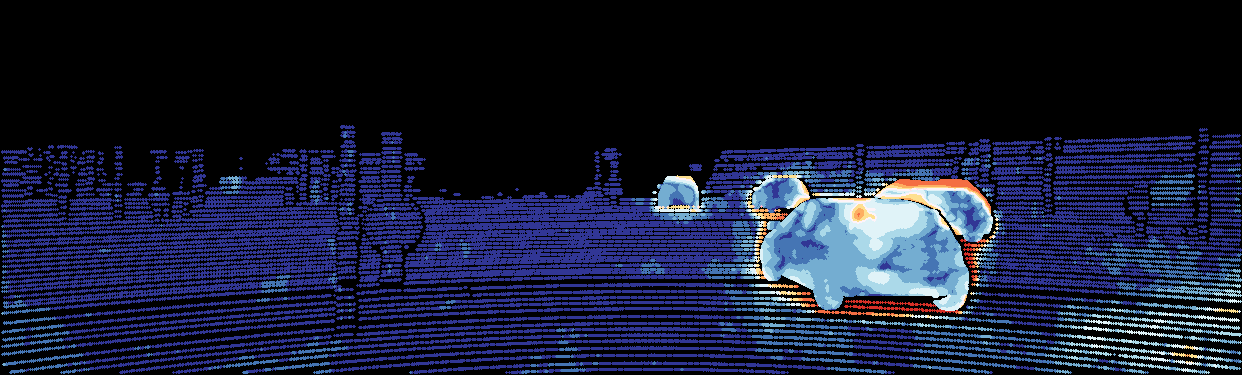}\\
    \includegraphics[width=3.4cm]{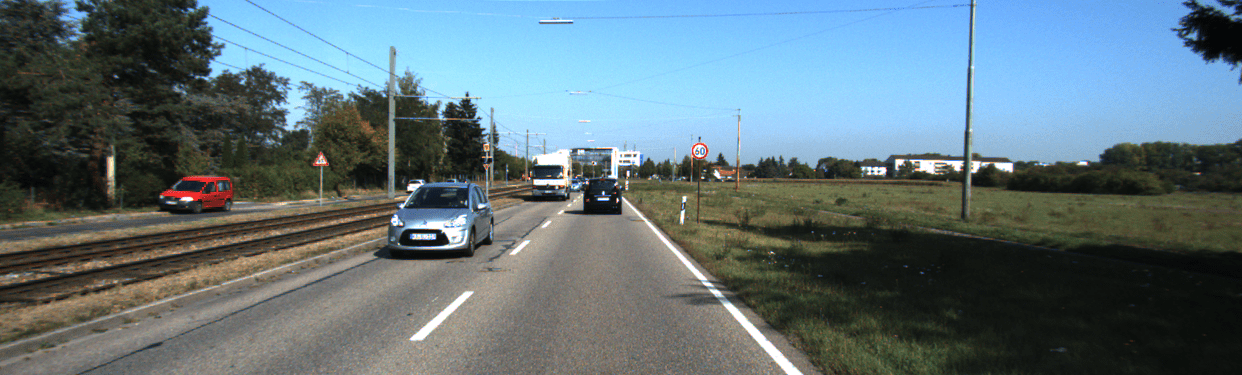}  &
    
    \includegraphics[width=3.4cm]{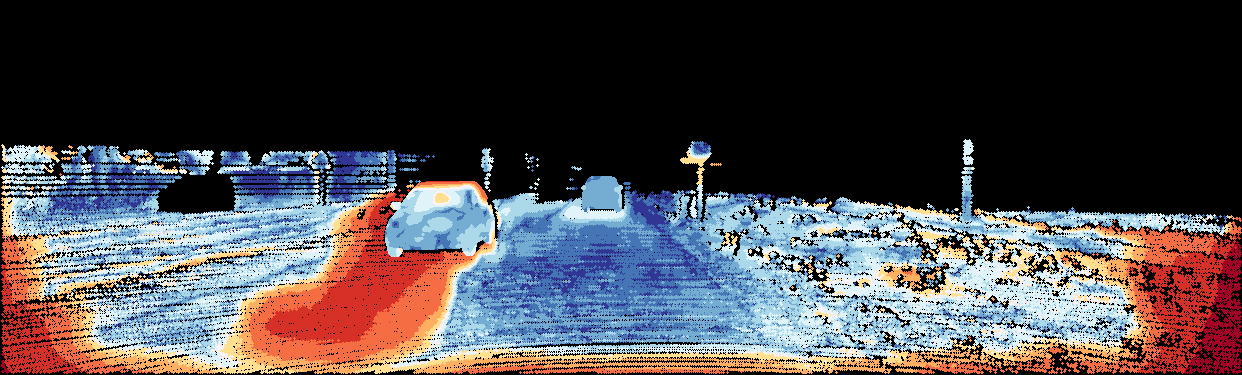} &
        \includegraphics[width=3.4cm]{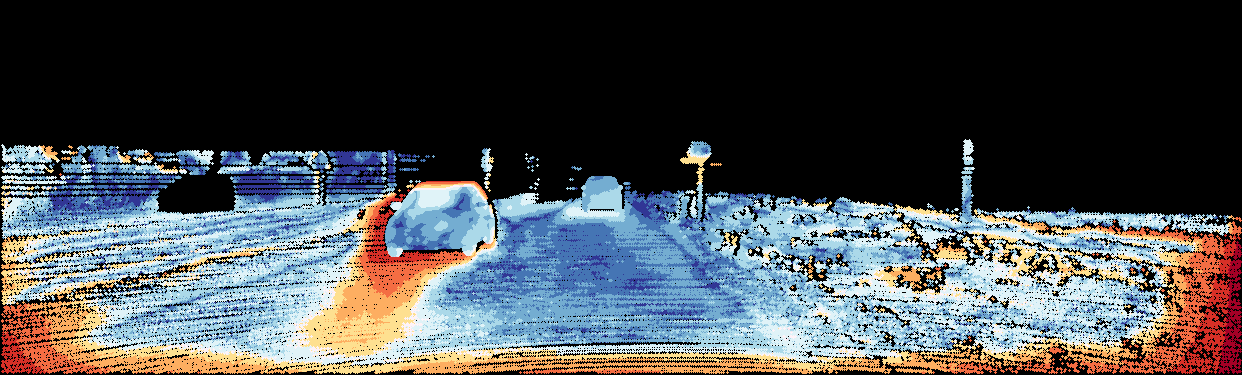} &
    \includegraphics[width=3.4cm]{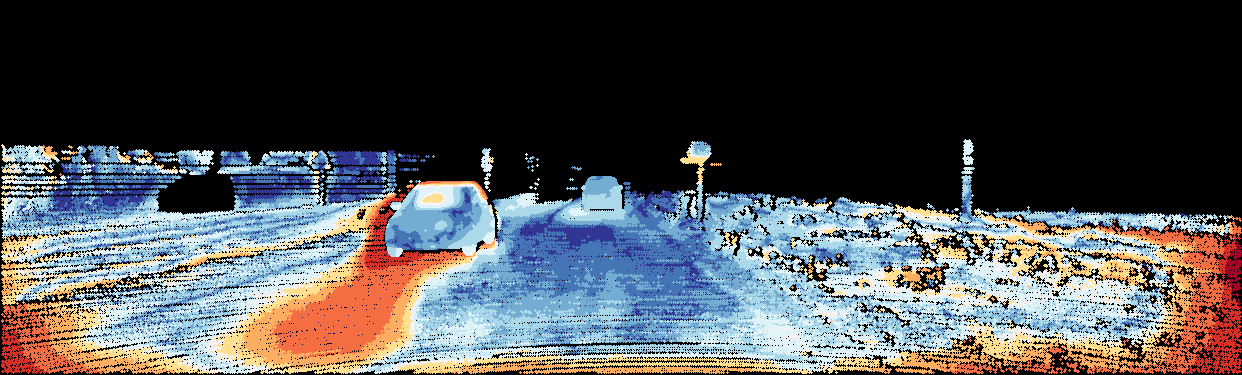} &
    \includegraphics[width=3.4cm]{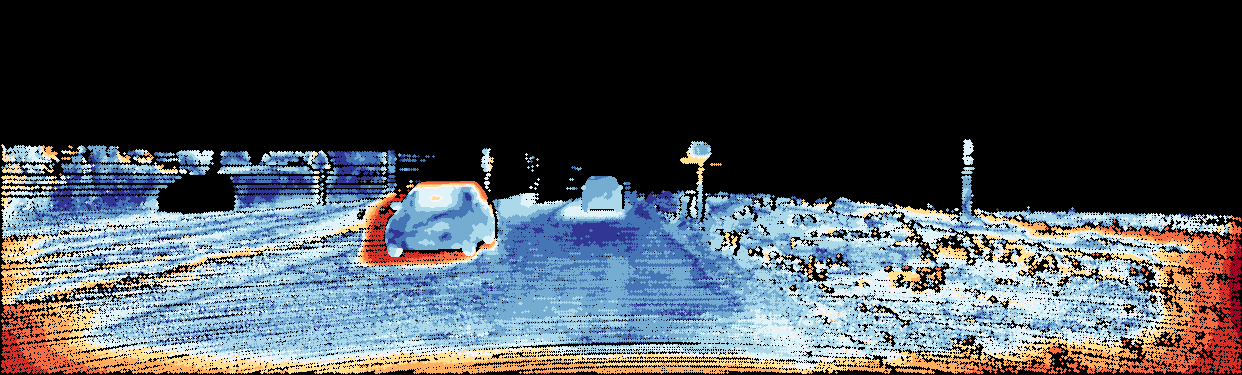}\\
    \includegraphics[width=3.4cm]{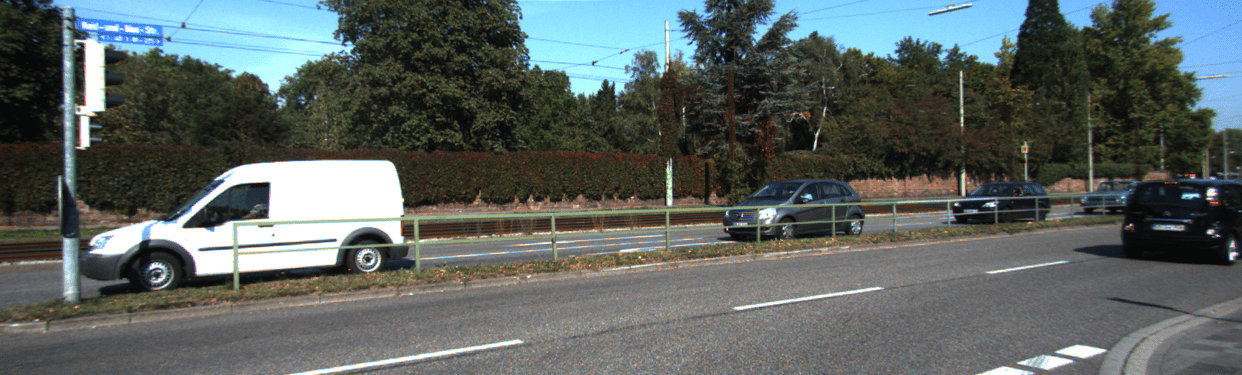}  &
    \includegraphics[width=3.4cm]{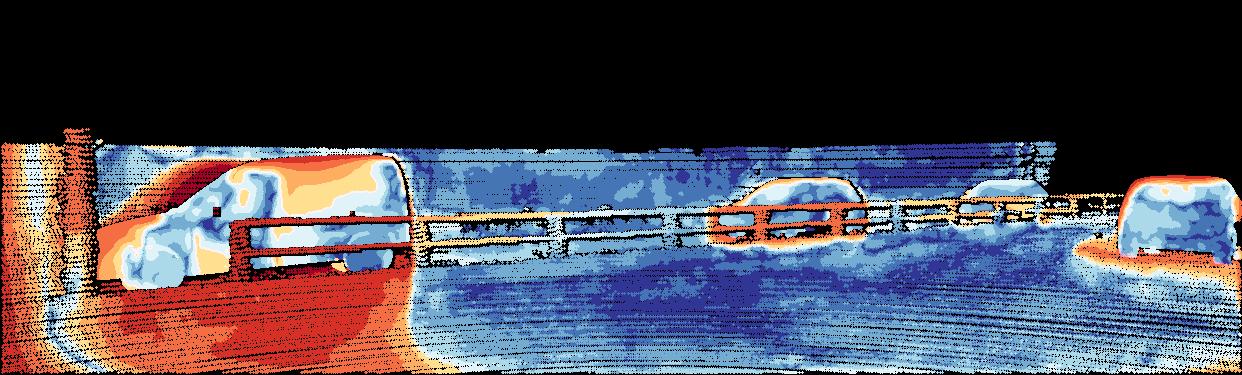} &
        \includegraphics[width=3.4cm]{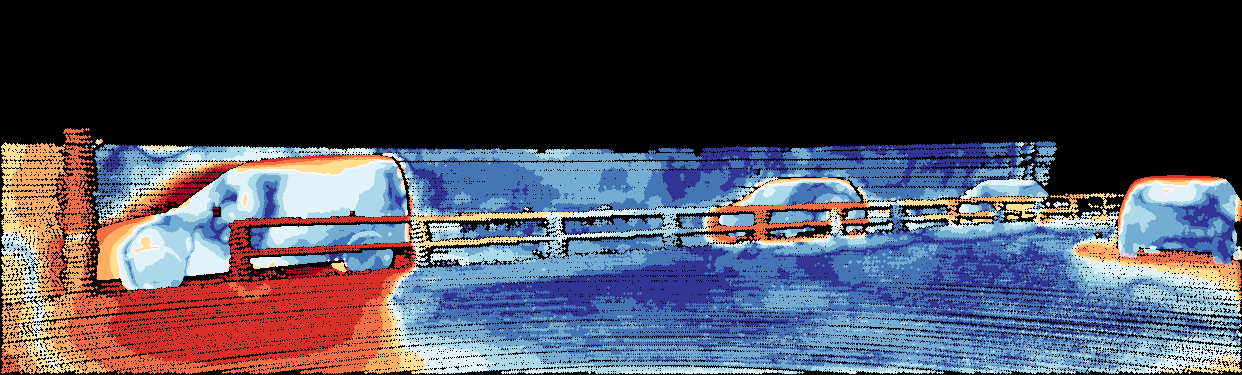} &
    \includegraphics[width=3.4cm]{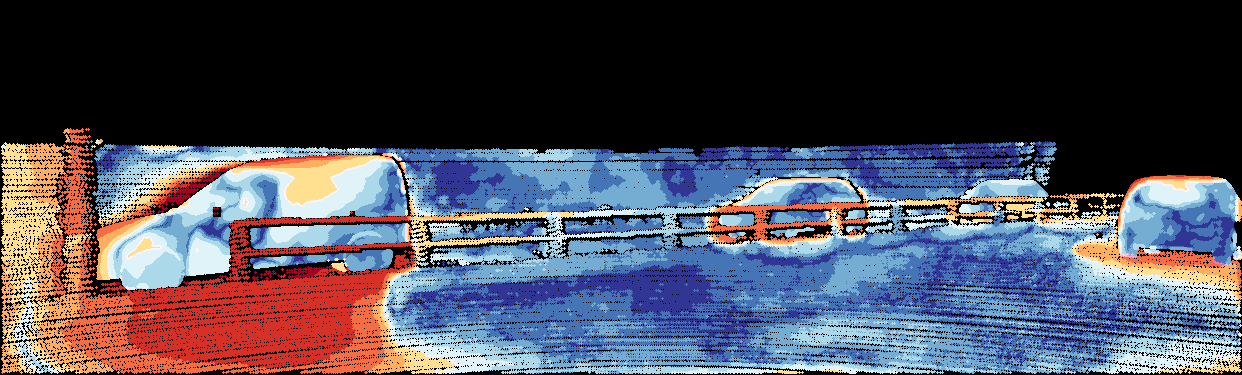} &
    \includegraphics[width=3.4cm]{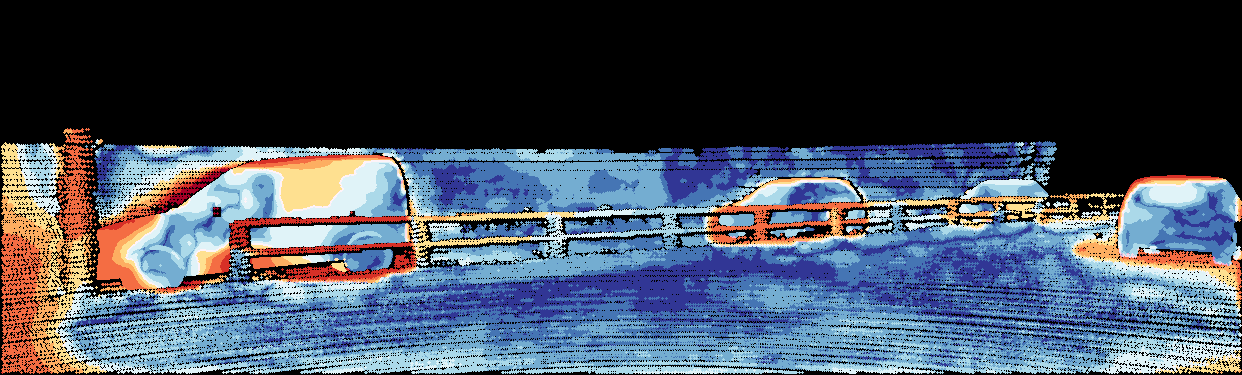}\\
    \end{tabular}
	\caption{\textbf{Endpoint error performance of our various models on the KITTI 2015 training dataset.} We compared Our-baseline, Our-F, Our-low-rank, and Our-sub models on the KITTI 2015 dataset to analyze their performance when handling dynamic objects. The results of the Our-sub model are much better.}
    \label{fig:comparison}
    \vspace{-0.2cm}
\end{figure*}
\vspace{-2mm}
\paragraph{KITTI VO training results.}
We report our results that were trained on the KITTI VO dataset in Table ~\ref{tab:results}, where our models are compared with various state-of-the-art methods. Our methods outperform all previous learning-based unsupervised optical flow methods with a notable margin. Note that most scenes in KITTI VO dataset are stationary, and therefore the difference between our-gtF, our-F, our-low-rank and our-sub is small across these benchmarks. 

\paragraph{Benchmark Fine-tuning Results.}
We fine-tuned our models on each benchmark and report the results with a suffix '-ft' in Table \ref{tab:ftresults}. For example, simply following the same hyper-parameters as before, we finetuned our models on the KITTI 2015 testing data. After fine-tuning, Our-sub model shows great performance improvement and achieved an EPE of 2.61 and 5.56 respectively on the KITTI 2012 and KITTI 2015 training datasets, which outperforms all the deep unsupervised methods and many supervised methods. Similarly, on the MPI Sintel trainings dataset, Our-sub-ft model performs best among the unsupervised methods, with an EPE of 3.94 on the Clean images and 5.08 on the Final images. Furthermore, both on the KITTI and Sintel testing benchmarks, our method outperformed the current state-of-the-art unsupervised method Back2Future Flow by a margin. We improve the best unsupervised performance from an Fl of 22.94\% to 16.24\% on KITTI 2015. The Our-sub-ft model achieved an EPE of 6.84 on the Sintel Clean dataset and 8.33 on the Final set, which are the results that unsupervised methods have never touched before. Additionally, it should be noted that the Back2Future Flow method is based on a multi-frame formulation while our method only requires two frames. Our model is also competitive compared with some fine-tuned supervised networks, such as SpyNet.   

Qualitatively, as shown in Fig.~\ref{fig:kitti_results} and Fig.~\ref{fig:sintel_results}, compared with the results of Back2Future Flow, the shapes in our estimated flows are more structured and have more explicit boundaries which represent motion discontinuities. This trend is also apparent in the flow error images. For example, on the KITTI 2015 dataset (Fig.~\ref{fig:kitti_results}), the results of Back2Future Flow usually bring a larger region of error with crimson colors around the objects.

It should be noted that fine-tuning on the target datasets (\eg, KITTI 2015) does not bring significant improvement because we have trained the models on a real-world dataset KITTI VO. The models have learned the general concepts of realistic optical flows and fine tuning just helps them familiar with the datasets' characteristics. On the KITTI 2012 training set, the fine-tuned model achieves very close results with the Our-sub model, which are respectively 2.61 and 2.62 EPE. Fine-tuning on the Sintel Clean dataset improves the result from 6.15 to 3.94 EPE, because the Sintel Clean dataset renders the synthetic scenes under low complexity and the images are quite different from the real world.

\subsection{Ablation study}

\begin{figure}[htbp]
\vspace{-0.2cm}
\centering
    \includegraphics[width=6cm]{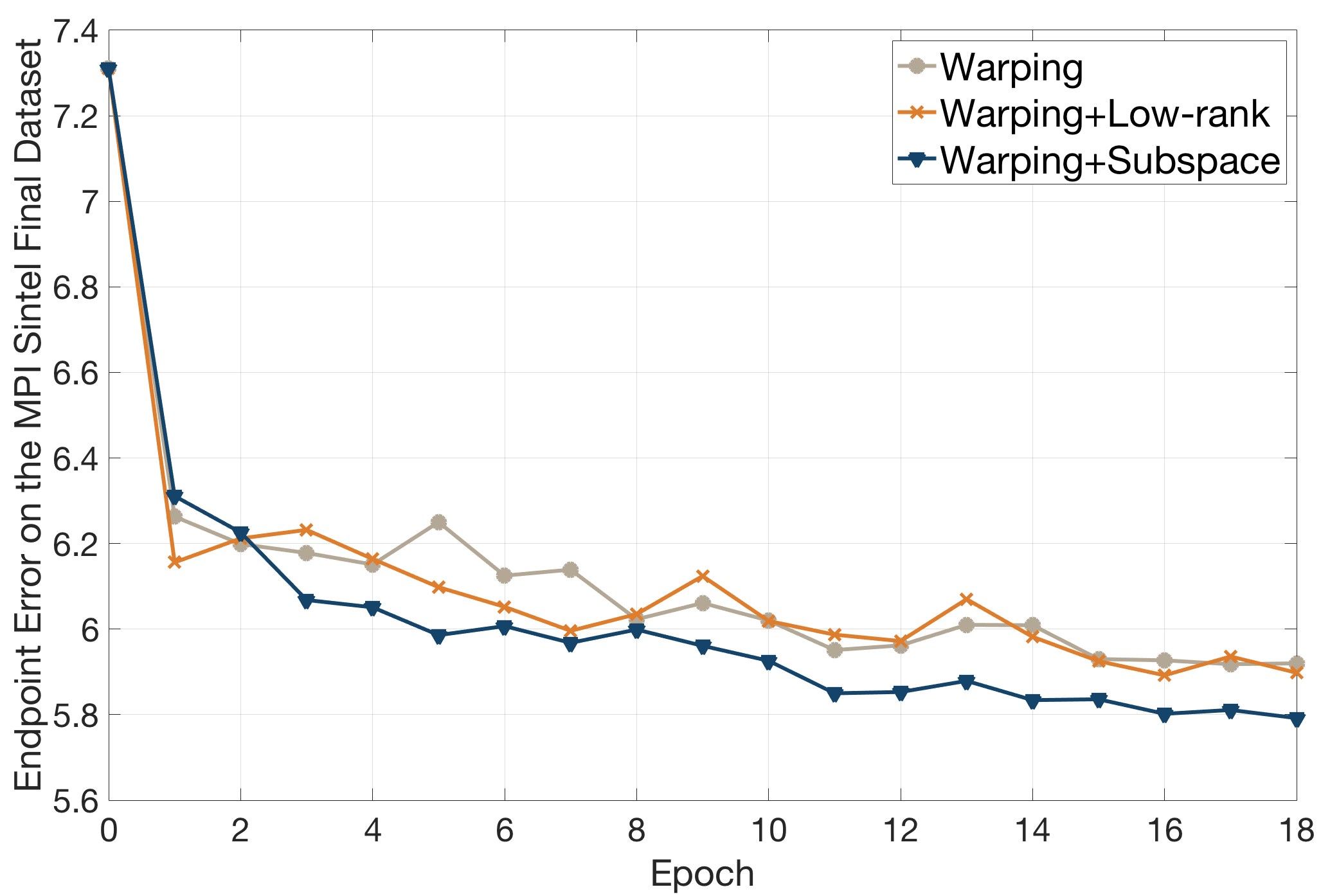}  
	\caption{\textbf{Endpoint error over epochs on the Sintel Final dataset.} We illustrate the endpoint errors over the training epochs when using various combinations of constraints. For all the three methods, the training started from the same pre-trained model `Our-baseline'. Combing the image warping and subspace constraints outperforms other two methods, which is consistent with the final fine-tuned results reported in Table \ref{tab:ftresults}.}
    \label{fig:valfig}
    \vspace{-0.3cm}
\end{figure}
\begin{table}[hbpt]
\setlength{\tabcolsep}{1.4mm}
	\centering
	\begin{tabular}{l l  c c c}
    \cmidrule(lr){1-5}
&Method & \multicolumn{2}{c}{KITTI 2015} & Sintel Final\\
\cmidrule(lr){3-4}
\cmidrule(lr){5-5}
& & EPE(all) &EPE(noc) & EPE(all)\\
\cmidrule(lr){1-5}
&Our-baseline-ft & 6.16 &2.85 & 5.87\\
&Our-F-ft & 6.19 &2.85 & NaN\\
&Our-low-rank-ft & 5.72 &2.62& 5.59\\
&Our-sub-ft & 5.56 &2.56 & 5.08\\
    \cmidrule(lr){1-5}
	\end{tabular}
    \caption{\textbf{Fine-tuning results comparison on KITTI 2015 and Sintel Final training sets.} We fine-tuned our models on the training sets of KITTI 2015 and Sintel Final dataset. The term NaN indicates the model cannot converge.}
\label{tab:ftresults}
\vspace{-0.3cm}
\end{table}

The Our-F, Our-low-rank and Our-sub models all work well in stationary scenes and they have similar quantitative performance. To further analyze their capabilities in handling general dynamic scenarios, we fine-tuned each method on the KITTI 2015 and Sintel Final dataset. Both of them  involve multiple motions in an image while Sintel scenes are more dynamic. As shown in Table \ref{tab:ftresults}, Our-sub can handle dynamic scenarios best and achieves the lowest EPE in both benchmarks. The hard fundamental constraint shares a similar performance with our baseline model but cannot converge on the Sintal dataset, whose EPE is reported as NaN. It is because a highly dynamic scene does not have a global fundamental $\mathbf{F}$. For the low-rank constraint, its performance is not affected by dynamic objects while it cannot gain information by modeling multiple movements as well. In Fig.~\ref{fig:valfig}, we provide the validation error curves over the training's early stages on Sintal final dataset. The subspace loss helps the model converge quicker and achieve lower cost than other methods.

\section{Conclusion}
In this paper, we have proposed effective methods to enforce global epipolar geometry constraints for unsupervised optical flow learning. For a stationary scene, we applied the low-rank constraint to regularize a globally rigid structure. For general dynamic scenes (multi-body or deformable), we proposed to use the union-of-subspaces constraint. Experiments on various benchmarking datasets have proved the efficacy and superiority of our methods compared with state-of-the-art (unsupervised) deep flow methods. In the future, we plan to study the multi-frame extension, \ie, enforcing geometric constraints across multiple frames.

\noindent
\textbf{Acknowledgement}
Yiran Zhong's research was supported in part by Australia Centre for Robotic Vision and Data61 CSIRO. Yuchao Dai's research was supported in part by the Natural Science Foundation of China grants (61871325, 61420106007). Hongdong Li's research was supported in part by the Australian Research Council (ARC) grants (LE190100080, CE140100016, DP190102261).

{\small
\bibliographystyle{ieee}
\bibliography{Unsupervised-Flow-Reference.bib}
}

\end{document}